\documentclass[10pt,twocolumn,letterpaper]{article}

\usepackage{cvpr}              

\usepackage{graphicx}
\usepackage{amsmath}
\usepackage{amssymb}
\usepackage{booktabs}
\usepackage{multirow}
\usepackage[ruled, lined]{algorithm2e}

\usepackage[pagebackref,breaklinks,colorlinks]{hyperref}

\usepackage[table,xcdraw]{xcolor}

\usepackage[accsupp]{axessibility}

\usepackage[capitalize]{cleveref}
\crefname{section}{Sec.}{Secs.}
\Crefname{section}{Section}{Sections}
\Crefname{table}{Table}{Tables}
\crefname{table}{Tab.}{Tabs.}

\newcommand{\category}[1]{{\texttt{#1}}}

\definecolor{objectColor}{rgb}{0.6, 0, 0}
\newcommand{\object}[1]{{\color{objectColor}{#1}}}
\definecolor{relationshipColor}{rgb}{0.415, 0.658, 0.309}
\newcommand{\relationship}[1]{{\color{relationshipColor}{#1}}}
\definecolor{actionColor}{rgb}{0.901, 0.568, 0.219}
\newcommand{\action}[1]{{\color{actionColor}{#1}}}
\definecolor{timeColor}{rgb}{0.403, 0.305, 0.654}
\newcommand{\temporal}[1]{{\color{timeColor}{#1}}}
\definecolor{conj}{rgb}{0, 0.5, 0.6}
\newcommand{\conj}[1]{{\color{conj}{#1}}}

\def\cvprPaperID{11269} 
\def\confName{CVPR}
\def\confYear{2022}

\begin{document}

\title{
Measuring Compositional Consistency for Video Question Answering
}

\author{Mona Gandhi$^1$*, Mustafa Omer Gul$^2$*, Eva Prakash$^2$, Madeleine Grunde-McLaughlin$^3$, \\ Ranjay Krishna$^3$, Maneesh Agrawala$^2$\\
Veermata Jijabai Technological Institute$^1$, Stanford University$^2$, University of Washington$^3$\\
{\tt\small \{mbgandhi\_b18\}@ce.vjti.ac.in, \{momergul, eprakash, maneesh\}@stanford.edu,} \\ {\tt\small \{mgrunde, ranjaykrishna\}@cs.washington.edu} \\
}

\maketitle

\def\thefootnote{*}\footnotetext{Equal contribution}\def\thefootnote{\arabic{footnote}}

\begin{abstract}
Recent video question answering benchmarks indicate that state-of-the-art models struggle to answer compositional questions.
However, it remains unclear which types of compositional reasoning cause models to mispredict. Furthermore, it is difficult to discern whether models arrive at answers using compositional reasoning or by leveraging data biases.
In this paper, we develop a question decomposition engine that programmatically deconstructs a compositional question into a directed acyclic graph of sub-questions. The graph is designed such that each parent question is a composition of its children. We present AGQA-Decomp, a benchmark containing $2.3M$ question graphs, with an average of $11.49$ sub-questions per graph, and $4.55M$ total new sub-questions. Using question graphs, we evaluate three state-of-the-art models with a suite of novel compositional consistency metrics. We find that models either cannot reason correctly through most compositions or are reliant on incorrect reasoning to reach answers, frequently contradicting themselves or achieving high accuracies when failing at intermediate reasoning steps.
\end{abstract}


\section{Introduction}
\begin{figure}[t]
    \centering
    \includegraphics[width=\columnwidth]{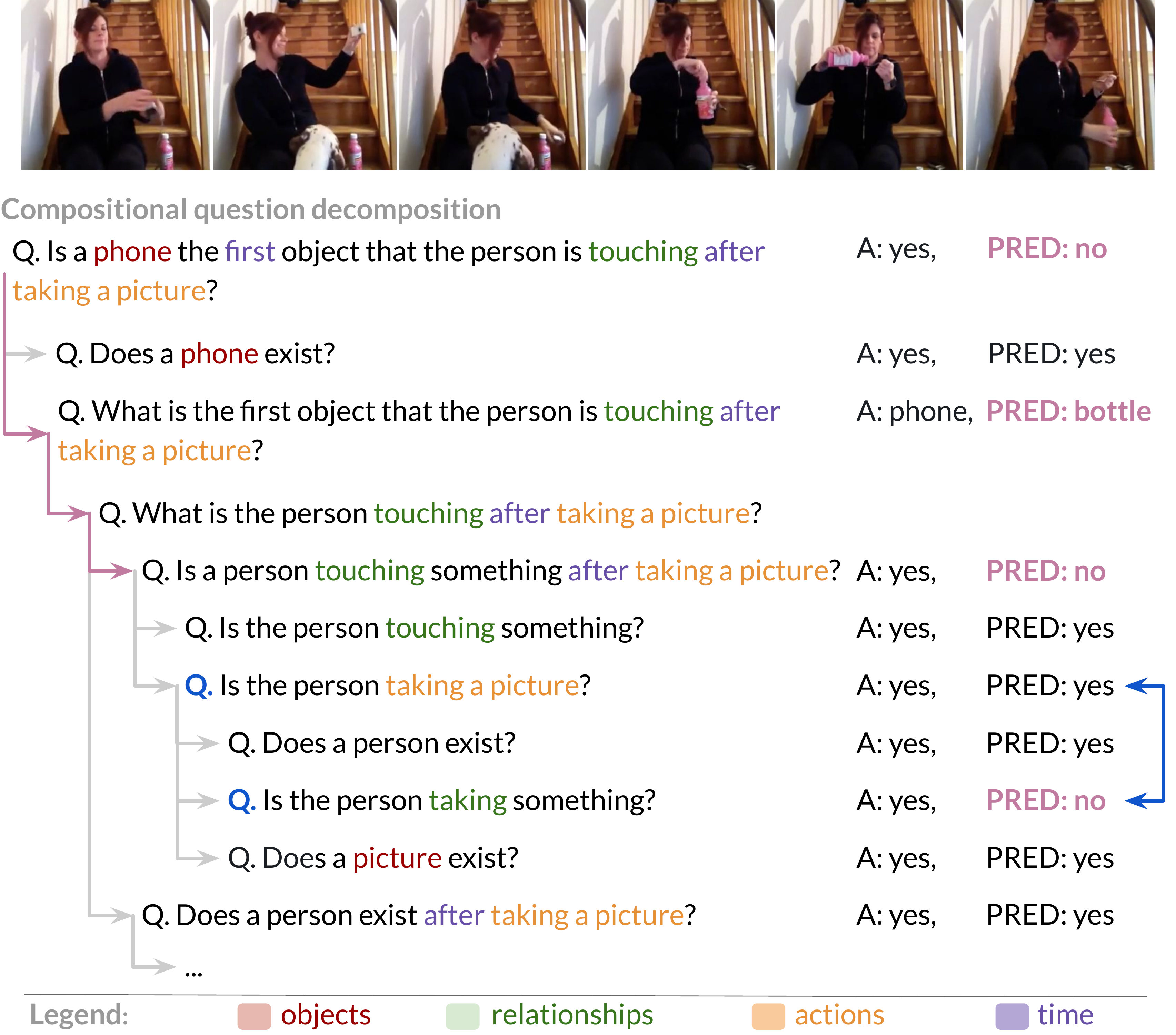}
    \caption{We introduce a question decomposition engine, which produces a DAG of sub-questions from a compositional question about visual events. A sub-question is designed to contain a subset of the original question's reasoning steps. Our engine produces a benchmark with $4.55M$ question answer pairs associated with $9.6K$ videos. We design handcrafted programs and templates for each sub-question as well as composition rules to compose sub-questions together. We analyze existing models using a suite of new compositional consistency metrics using our DAGs. Our DAGs isolate which composition rules cause mispredictions (error path is shown by pink arrows). They also highlight scenarios where models might exhibit self-contradiction (blue arrows).}
    \label{fig:pull}
\end{figure}

Compositional reasoning is fundamental to how humans represent visual events ~\cite{reynolds2007computational,speer2007human,kurby2008segmentation,lillo2014discriminative}. 
For instance, Figure~\ref{fig:pull} visualizes a video consisting of actions such as \action{taking a picture} and \action{holding a bottle}; the action \action{holding a bottle} involves an actor initially \relationship{twisting} the \object{bottle} and then later \relationship{holding} it.
This ability to compose interactions and actions is reflected in the compositional nature of language people use to communicate about what they see~\cite{chomsky2002syntactic,montague1970universal}. To measure compositional reasoning of visual events, the computer vision community has proposed multiple video benchmarks using question answering~\cite{lei2018tvqa,tapaswi2016movieqa,GrundeMcLaughlin2021AGQA}. These benchmarks ask questions such as ``Is a \object{phone} the \temporal{first} object that the person is \relationship{touching} \temporal{after} \action{taking a picture}?'', where models need to compose actions (\action{taking a picture}) with relationships (\relationship{touching}) and objects (\object{phone}) to arrive at the correct answer.
Using these benchmarks, researchers have recently concluded that state-of-the-art models~\cite{fan2019heterogeneous,le2020hierarchical, li2019beyond} struggle to reason compositionally~\cite{GrundeMcLaughlin2021AGQA}.

\begin{table*}[]
\caption{We visualize our hand-designed sub-questions, which consist of a subset of the reasoning steps found in the AGQA benchmark~\cite{GrundeMcLaughlin2021AGQA}. Each sub-question consists of a functional program and a natural language template.}
\label{tab:subquestiontypes}
\centering
\resizebox{\linewidth}{!}{
\begin{tabular}{lll}

\textbf{Sub-question type}                                                                                                                                                & \textbf{Description}                                                                                                                                                                 & \textbf{Example}                                                                                                                           \\ \hline
\category{Object exists}                                                                                                                                                   & To verify if an \object{object} exists                                                                                                                                           & Does a \object{doorway} exist?                                                                                                             \\ \hline
\category{Relation exists}                                                                                                                                                 & To verify if a \relationship{relationship} exists                                                                                                                                          & Is the \object{person} \relationship{holding} something?                                                                                                  \\ \hline
\category{Interaction}                                                                                                                                                     & \begin{tabular}[c]{@{}l@{}}To verify if there is a particular \relationship{relationship} \temporal{between} \\ \object{person} and an \object{object}\end{tabular}                                                        & Is the \object{person} \relationship{touching} a \object{dish}?                                                                                                     \\ \hline
\category{Interaction temporal loc.}
& A filter on an interaction type question                                                                                                                               & \begin{tabular}[c]{@{}l@{}}Is the \object{person} holding a book \\ \temporal{while} \action{smiling at something}?\end{tabular}                               \\ \hline
\category{Exists temporal loc.}                         & A condition on \object{object}/\relationship{relationship} exists question                                                                                                                          & Does a \object{phone} exist \temporal{after} \action{looking in the mirror}?                                                                                   \\ \hline

\category{First/last} & Getting the first/last instance of the given \object{object}                                                                                                                             & \begin{tabular}[c]{@{}l@{}}What is the \temporal{first} object that the \object{person} is \relationship{above} \\ \temporal{before} \action{walking through the doorway}?\end{tabular}  \\ \hline

\category{Longest shortest action}                                                                                                                                       & Getting the \temporal{longest}/\temporal{shortest} \action{action}                                                                                                                                     & \begin{tabular}[c]{@{}l@{}}What does the \object{person} do for \\ the \temporal{longest} amount of time?\end{tabular}                                \\ \hline

\category{Conjunction} & \begin{tabular}[c]{@{}l@{}}Get a new exists question by combining two \\ interaction questions with a conjunction\end{tabular}                                          & \begin{tabular}[c]{@{}l@{}}Is the \object{person} \relationship{in front of} the \object{mirror} \conj{and} \relationship{behind} \\ the \object{table} \temporal{while} \action{looking in the mirror}?\end{tabular} \\ \hline
\category{Choose} & \begin{tabular}[c]{@{}l@{}}Compares between two \object{objects}, \action{actions},\\ \relationship{relationships}, or \temporal{time lengths}   \end{tabular}                                                                                                                                 & \begin{tabular}[c]{@{}l@{}}Is the \object{doorknob} or the \object{dish} the \temporal{first} object that \\ the \object{person} is \relationship{holding}?\end{tabular}                \\ \hline
\category{Equals} & \begin{tabular}[c]{@{}l@{}}Compares two \object{objects} and verifies if they are the same \\ Verifies if the given \action{action} is \temporal{longer}/\temporal{shorter} \\ than the other one\end{tabular} & \begin{tabular}[c]{@{}l@{}}Is the \object{doorway} the object they are interacting with \\ \temporal{while} \action{holding a dish}?\end{tabular}               \\ \hline
\end{tabular}
}
\end{table*}

Unfortunately, existing benchmarks are unable to explain \textit{why} video question answering models struggle with compositional reasoning. In Figure~\ref{fig:pull}, a model incorrectly answers the root question as ``no'' instead of the correct answer of ``yes.''
However, this information does not explain what caused the model to err: 
Did the model struggle with words requiring temporal reasoning, such as \temporal{first} or \temporal{after}? Did it fail at detecting the \object{phone} or identifying the relationship \relationship{touching}? Or did it struggle to compose the relationship with the object? Even if we assume the model had correctly answered the question, it remains uncertain whether this behavior was due to proper compositional reasoning or a reliance on spurious correlations to ``cheat.''

Not only do standard evaluation schemes fall short in this regard, but existing approaches for dissecting model behavior also struggle to resolve this uncertainty. Attribution methods, such as GradCAM~\cite{selvaraju2017grad} or LIME~\cite{ribeiro2016should}, can highlight important aspects of the input data, but are agnostic to the structure of compositional reasoning. Approaches that rely on counterfactuals to illuminate model behavior, such as contrast sets~\cite{gardner2020evaluating}, focus primarily on model decision boundaries by performing minor, local changes to the input. These local changes, however, cannot capture the full range of compositional reasoning steps required to answer compositional visual questions~\cite{GrundeMcLaughlin2021AGQA}, which assess multiple, often interdependent, reasoning abilities at once.

In this paper, we develop a question decomposition engine that decomposes a compositional question into a directed acyclic graph (DAG) of sub-questions (see Figure~\ref{fig:pull}). A sub-question isolates a subset of the reasoning steps that the original question requires, exposing model performance on subsets of intermediate reasoning steps. This exposure enables us to identify difficult sub-questions and study which compositions cause models to struggle. It also allows us to test whether models are right for the right reasons. For instance, the root question mentioned earlier can not only decompose into intermediate reasoning steps that determine if the ``the \object{person} was \relationship{touching} something \temporal{after} \action{taking a picture},'' but also isolate basic perception capabilities, such as determining whether a ``\object{phone} exists''.

Using our engine, we construct the AGQA-Decomp dataset\footnote{Project page: \url{https://tinyurl.com/agqa-decomp}}, which decomposes the $2.3M$ compositional questions in the updated version\footnote{AGQA 2.0: \url{https://tinyurl.com/agqavideo}} of the recent balanced AGQA benchmark~\cite{GrundeMcLaughlin2021AGQA} to produce $1.62M$ unique sub-questions for $9.6K$ videos for a total of $4.55M$ sub-questions.
To generate sub-questions, we hand-design $21$ sub-questions, each with a functional program and natural language template (Table~\ref{tab:subquestiontypes}). To compose the sub-questions within a DAG, we hand-design $13$ composition rules (Table~\ref{tab:compositiontypes}).
Finally, we create a suite of new metrics to evaluate compositional reasoning. One of those metrics --- internal consistency --- measures whether models are self-consistent when they answer questions within a DAG. To enable this metric, we further hand-design $10$ consistency rules between sub-questions (see Table~\ref{tab:consistencyrules} in the Supplementary).

\begin{figure*}[t]
    \centering
    \includegraphics[width=1\linewidth]{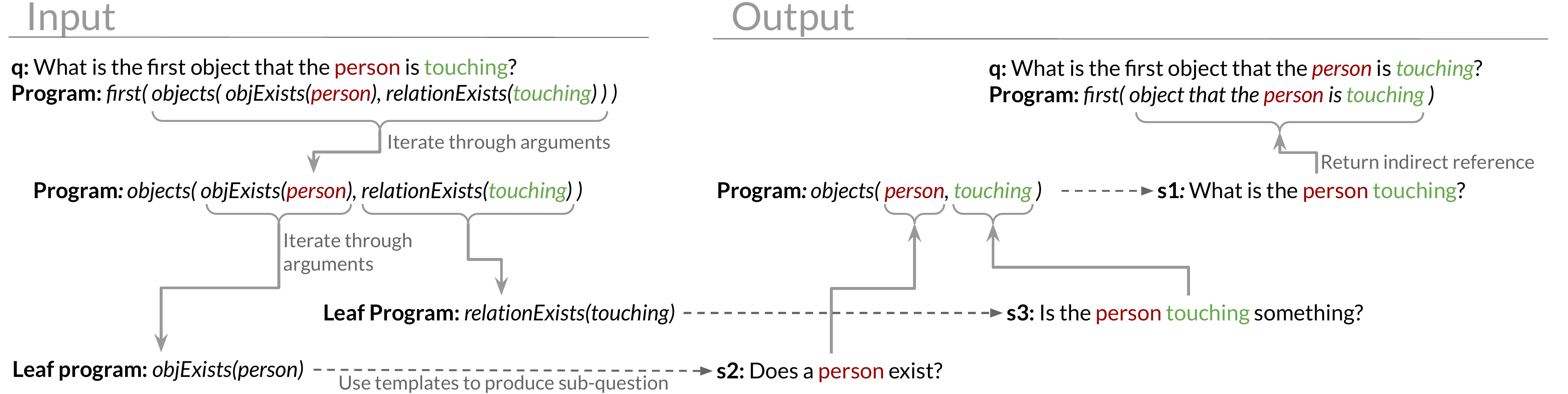}
    \caption[width=1\linewidth]{Our question decomposition engine expects a compositional root question as input and outputs a DAG of sub-questions. The root question has an associated functional program which explains the reasoning steps necessary to answer the question. We recursively iterate over the arguments of the function until we reach a leaf function. We design natural language templates for each leaf function, converting them into sub-questions. Once a leaf function is converted to a question, we return an indirect reference of the answer back to its parent. The parent uses composition rules to combine the indirect references from its children to similarly generate questions.
    }
    \label{fig:system}
\end{figure*}

We evaluate three state-of-the-art video question answering models, HCRN~\cite{le2020hierarchical}, HME~\cite{fan2019heterogeneous} and PSAC~\cite{li2019beyond} using our DAGs and metrics. Our analyses reveal that for a majority of compositional reasoning steps, models either fail to successfully complete the step or rely on faulty reasoning mechanisms. They frequently contradict themselves and achieve high accuracies even when failing at intermediate steps. Models even struggle when asked to choose between or compare two options, such as objects or relationships. Finally, we find that there is a weak negative correlation between internal consistency and accuracy across DAGs for each model. From the models we evaluated, HME obtains the most negative correlation, suggesting that the model is frequently inaccurate and propagates this inaccuracy due to its internal consistency. We believe that our decomposed question DAGs could further enable a host of future research directions: from promoting transparency through consistency to developing interactive model analysis tools.

\section{Related Work}

We contrast our contributions against recently proposed evaluation measures in machine learning, focusing especially on video question answering. We also contextualize the idea of question decomposition to related work in computer vision and in natural language processing (NLP).

\noindent\textbf{Video question answering.} Despite the popularity of video question answering as a benchmark task~\cite{tapaswi2016movieqa,jang2017tgif,lei2018tvqa,yu2019activitynet,Girdhar2020CATER:,Yi*2020CLEVRER:,GrundeMcLaughlin2021AGQA}, questions in several prominent benchmarks rely on dialogue and plot summaries instead of a video's visual contents~\cite{lei2018tvqa,tapaswi2016movieqa,kim2017deepstory,zadeh2019social}, focus on short video clips or only a handful of objects~\cite{yi2019clevrer, mun2017marioqa}, or suffer from biases associated with human generated questions~\cite{yu2019activitynet,tapaswi2016movieqa,jang2017tgif,lei2018tvqa}. These limitations reduce benchmarks' effectiveness at reasoning over compositional visual events.
Given these limitations, we focus on the recent AGQA benchmark~\cite{GrundeMcLaughlin2021AGQA} of question answer pairs for compositional visual reasoning.

\noindent\textbf{Evaluating consistency.} Our focus on providing an evaluation metric beyond standard task accuracy is in line with recent efforts toward more metamorphic evaluation of machine learning models~\cite{gardner2020evaluating,bitton2021automatic,li2020linguistically}. While we may be the only method to date proposing a consistency-based metric for video question answering, the role of consistency has been explored for image question answering~\cite{ribeiro2019red,ray2019sunny,shah2019cycle,hudson2019gqa,selvaraju2020squinting,gokhale2020vqa,bitton2021automatic,yuan2021perception} and for text question answering~\cite{gardner2020evaluating,wu2021polyjuice}.
Existing metrics measure whether models can consistently answer sets of questions logically entailed by a given question~\cite{ribeiro2019red,ray2019sunny,hudson2019gqa,gokhale2020vqa} or answer counterfactuals with different answers~\cite{gardner2020evaluating,wu2021polyjuice}. To enable these metrics, researchers have collected datasets by asking human annotators to generate perceptual questions associated with reasoning questions~\cite{selvaraju2020squinting}, used large language models to generate counterfactuals~\cite{wu2021polyjuice}, or asked domain experts to compile rules to generate contrast sets~\cite{gardner2020evaluating}. In comparison, we programmatically decompose questions by hand-designing composition rules over programs associated with questions.

\noindent\textbf{Decomposing question answering.} Decomposing the question answering task into simpler tasks has appeared within both the computer vision~\cite{andreas2016neural,cao2018visual} and NLP communities~\cite{wolfson2020break}. Most prominently in computer vision, neural module networks and related architectures~\cite{andreas2016neural,hu2017learning,chen2021meta} break down questions into modular programs defining the architecture of the neural network instantiated to answer the question. To design modular architectures, ACMN~\cite{cao2018visual} decomposes questions using dependency parses. The GQA~\cite{hudson2019gqa} and AGQA~\cite{GrundeMcLaughlin2021AGQA} benchmarks use programs associated with each question to compute answers from scene graphs~\cite{krishna2017visual} and spatio-temporal scene graphs~\cite{ji2020action}; however, these programs are unused beyond dataset generation.

\begin{table*}[]
\caption{We hand-design composition rules to generate questions $q$ using indirect references produced by its sub-questions $\{s_1, s_2, \ldots\}$.}
\label{tab:compositiontypes}
\centering
\resizebox{\linewidth}{!}{
\begin{tabular}{lll}

\textbf{Composition rules}                                                                                 & \textbf{Description}                                                                                                           & \textbf{Example}                                                                                                                                                                                                                                                                                  \\ \hline
 \category{Interaction}                                                                                      & Verify if an interaction exists                                                                                   & \begin{tabular}[c]{@{}l@{}}\emph{q} : Is a \object{person} \relationship{holding} a \object{doorway}? \\ \emph{s1}:  Does a \object{person} exist? \\ \emph{s2}: Is a person \relationship{holding} something? \\ \emph{s3}:  Does a \object{doorway} exist?\end{tabular}                                                                                                               \\ \hline
\begin{tabular}[c]{@{}l@{}}\category{Temporal loc.} \\ \category{(After, before, while, between)}\end{tabular} & \begin{tabular}[c]{@{}l@{}}Combine two interaction or exists \\ questions using a temporal localizer\end{tabular} & \begin{tabular}[c]{@{}l@{}}\emph{q}: Is the person \action{touching a doorway} \temporal{before} \action{smiling at something}? \\ \emph{s1}: Is the person \action{touching a doorway}? \\ \emph{s2}: Is a person \action{smiling at something}?\end{tabular}                                                                                \\ \hline
\category{First/last}                                                                                       & \begin{tabular}[c]{@{}l@{}}Getting the first/last occurrence \\ from a set of object/actions\end{tabular}          & \begin{tabular}[c]{@{}l@{}}\emph{q}: What is the \temporal{first} object that the \object{person} is \relationship{holding}? \\ \emph{s1}: What is the \object{person} \relationship{holding}?\end{tabular}                                                                                                                                    \\ \hline
\begin{tabular}[c]{@{}l@{}}\category{Conjunction} \\ \category{(And, xor)}\end{tabular}                                & \begin{tabular}[c]{@{}l@{}}Combine two interaction questions \\ using a conjunction\end{tabular}                  & \begin{tabular}[c]{@{}l@{}}\emph{q}: Is the person \action{putting some clothes} \conj{and} \action{behind a book} \temporal{before} \action{walking through the doorway}? \\ \emph{s1}: Is the person \action{putting some clothes} \temporal{before} \action{walking through the doorway}? \\ \emph{s2}: Is the person \action{behind a book} \temporal{before} \action{walking through the doorway}?\end{tabular} \\ \hline
\begin{tabular}[c]{@{}l@{}}\category{Choose} \\ \category{(Choose (object/Time)}\\ \category{longer/shorter choose)}\end{tabular}                & Chooses one of two possible options                                                                               & \begin{tabular}[c]{@{}l@{}}\emph{q}: Is the \object{doorway} or the \object{book} the \temporal{first} object they were in front of? \\ \emph{s1}: Is the \object{doorway} the \temporal{first} object they were in front of? \\ \emph{s2}: Is the \object{book} the \temporal{first} object they were in front of?\end{tabular}                                                   \\ \hline
\category{Equals}                                                                                           & \begin{tabular}[c]{@{}l@{}}Compares two objects/actions to \\ verify if they are the same\end{tabular}            & \begin{tabular}[c]{@{}l@{}}\emph{q}: Is a \object{book} the \temporal{first} object that the \object{person} is \relationship{carrying}? \\ \emph{s1}: Does a \object{book} exist? \\ \emph{s2}: What is the \temporal{first} object that the \object{person} is \relationship{carrying}?\end{tabular}                                                                                                \\ \hline
\end{tabular}
}
\end{table*}

In NLP, ``multi-hop'' reasoning questions are decomposed into ``single-hop'' ones (e.g.~ decomposing ``Which team does the player named 2015 Diamond Head
Classic’s MVP play for?'' into the simpler ``Which player was named 2015 Diamond Head Classic’s MVP?''). Multi-hop models answer simpler questions and combine their answers to ultimately answer the original multi-hop question~\cite{min2019multi,perez2020unsupervised}. In a similar vein, explanation methods have decomposed language statements into tree-structured sets of premises that entail the original statement (e.g.~``eruptions block sunlight'' entails ``eruptions can kill plants'')~\cite{dalvi2021explaining}. 
While BreakItDown~\cite{wolfson2020break} decomposes questions for HotPotQA~\cite{yang2018hotpotqa} into programs to design neural architectures, we decompose questions to design evaluation metrics.

\noindent\textbf{Compositional reasoning.}
While multiple definitions of compositionality exist, we use what is more colloquially referred to as bottom-up compositionality --- ``the meaning of the whole is a function of the meanings of its parts''~\cite{cresswell1973logics}. In our case, reasoning about the question ``Was the person \action{holding a bottle} \temporal{after} \relationship{touching} a \object{phone}?'' entails being able to answer simpler questions (e.g.~``Did the person \relationship{touch} a \object{phone}?''), which can be further decomposed into perceptual questions (e.g.~, ``Does a \object{phone} exist?'') and spatio-temporal relationship detection (e.g.~``Did the person \relationship{touch} something?'').
Recent work has argued the importance of compositionality in enabling models to generalize to new domains, categories, and logical rules~\cite{lake2018generalization,vatashsky2020vqa} and has discovered that current models struggle with multi-step reasoning~\cite{fan2019heterogeneous,GrundeMcLaughlin2021AGQA}. These studies motivate our contribution.

\section{Question decomposition engine}

Given a question $q$ as input, our engine outputs a directed acyclic graph (DAG) $(N_q, E_q) \in G_q$ of sub-questions for that question. The nodes $N_q$ are the list of sub-questions for question $q$ while the directed edges identify the composition rule used to compose a question from a node's sub-questions. For example, the decomposition of ``What is the \temporal{first} object that the \object{person} is \relationship{touching}?'' will produce the following list of sub-questions: $\{s1$: ``What is the \object{person} \relationship{touching}?'', $s2$: ``Does a \object{person} exist?'', and $s3:$ ``Is the \object{person} \relationship{touching} something?'' $\}$. The edges are: $\{(q, s1, \textrm{first}), (s1, s2, \textrm{interaction}), (s1, s3, \textrm{interaction})\}$, where ``first'' and ``interaction'' are composition rules.

To generate the DAG, we first represent the question $q$ as a functional program, which consists of the individual reasoning steps needed to answer $q$. The program structure defines the structure of the DAG (as shown in Figure~\ref{fig:system}). We recursively iterate over this program and its arguments to generate the DAG.

While our composition rules and templates are tailored towards AGQA~\cite{GrundeMcLaughlin2021AGQA}, our engine can be generalized to other datasets involving questions paired with functional programs, such as GQA~\cite{hudson2019gqa}, CLEVR~\cite{johnson2017clevr} or CLEVRER~\cite{yi2019clevrer}. This will require defining composition rules and templates based on the datasets' function programs.

\subsection{Representing questions as programs}\label{sec:AGQA to new}

We assume all questions have a corresponding functional program, with multiple reasoning steps. For instance, the program for $q$ is \texttt{first(objects(objExists(person), relationExists(touching)))}. Intuitively, this particular program searches through all the frames of a given video to find instances where there is a \object{person} present: \texttt{objExists(person)}. Similarly, it finds the frames where a person is \relationship{touching} something: \texttt{relationExists(touching)}. From those frames, it extracts the objects that are being \relationship{touched} by a \object{person}: \texttt{objects(objExists(person), relationExists(touching))}. Finally, it returns the \temporal{first} object from the list of objects identified: \texttt{first($\cdot$)}.

Each reasoning step is a function composed of multiple arguments: For example, the function \texttt{objects($\cdot$)} contains the following arguments: \texttt{objExists($\cdot$)} and \texttt{relationExists($\cdot$)}. 
We utilize the $2.3M$ questions, each generated using $27$ unique functions associated with $217$ natural language templates, in AGQA.

\subsection{Decomposing questions using programs}\label{sec:DAG}

To decompose $q$, we topologically iterate over all the arguments of the top-level reasoning function and recursively decompose each argument. For instance, the top level reasoning function for $q$ is \texttt{first($\cdot$)}. We iterate over its argument \texttt{objects($\cdot$)} and then recursively iterate over its two arguments: \texttt{objExists($\cdot$)} and \texttt{relationExists($\cdot$)}. 

Eventually, we will arrive at a ``leaf'' program with no further functions as arguments (e.g. \texttt{objExists(person)}). To convert the leaf program into a node in the DAG, we design natural language question templates for every program (see Table~\ref{tab:subquestiontypes}). For instance, \texttt{objExists($\cdot$)} has the template: ``Does an [\object{object}] exist?'' that creates the subquestion $s2$. We check if we have already added $s2 \in N_q$ while traversing another argument. If $s2 \notin N_q$, then we use the template to create a new node $s2=$ \textrm{``Does a \object{person} exist?''} and add it to $N_q$. 

Once we convert a leaf function into $s2$, we parse the template to extract an indirect reference and send it back to its parent function. The parent function, in this case \texttt{objects(objExists(person), relationExists(touching))} uses its arguments $s2$ and $s3$, along with a compositionality rule to produce the node $s1=$ ``What is the \object{person} \relationship{touching}?''. We design a set of compositionality rules, listed in Table~\ref{tab:compositiontypes}, to ingest the indirect references passed back ($s2 \rightarrow$ ``person'' and $s3 \rightarrow$ ``touching'') into its corresponding template: ````What is the [\object{object}] [\relationship{relationship}]?''. Next, we add the edges between $s1$ and its two arguments to $E_q$ with the composition rule, \category{interaction}, used to compose the arguments together.
This process continues until we return back to the original top-level function \texttt{first($\cdot$)}.

Our recursive decomposition process makes an average of $11.49$ sub-questions for each of the $2.3M$ questions in the balanced AGQA questions, creating $4.55M$ sub-questions. 

\subsection{AGQA answer generation}\label{sec:answer generation}

Once all the questions are decomposed into DAGs of sub-questions, we programmatically propagate answers from the original AGQA questions to the sub-questions. Some sub-questions are already present in the original unbalanced AGQA dataset; for these, we automatically have the answers. For others, we craft logical consistency rules to generate answers (see Table~\ref{tab:consistencyrules} in the Supplementary).

For example, if the answer to an \category{Interaction} question is ``yes'', then all its sub-questions should also be answered ``yes''. If the answer to ``Is the \object{person} \relationship{touching} something?'' is ``yes,'' for instance, then the answer to ``Does a \object{person} exist?'' is also ``yes''. If a ``choose X or Y'' question's answer is ``X'', then all sub-questions along X's recursive call should be answered ``yes,'' while Y's answer should be ``no.'' If, for example, ``Did the \object{person} \relationship{throw} the \object{blanket} but not \relationship{hold} the \object{blanket}?'' is answered ``yes'', then the answer to ``Did the \object{person} \relationship{throw} the \object{blanket}?'' is ``yes'' but ``Did the \object{person} \relationship{hold} the \object{blanket}'' is ``no''. Similar logical rules apply for \category{Before} and \category{After} question types.

Our answer generation rules are unable to propagate answers for questions answered ``no''. For instance, if the answer to ``Is the \object{person} \relationship{touching} something?'' is ``no'', we can not entail an answer to the question ``Does a \object{person} exist?''. 
To answer such questions, we run a large-scale annotation task on Amazon Mechanical Turk to identify all objects that appear in a randomly selected subset of videos in AGQA (see Supplementary for details). We use these annotations to propagate ``no'' answers to the relevant sub-questions.

\begin{table*}[]
\caption{We report accuracy, compositional accuracy (\textbf{CA}), right for the wrong reasons (\textbf{RWR}), delta (\textbf{RWR-CA}) and internal consistency (\textbf{IC}) values. We also present accuracy for the Most-Likely baseline and the rate at which annotators agreed with ground-truth answers in our AMT study (Human). Models particularly struggle at \category{Interaction Temporal Localization}, \category{Choose} and \category{Equals} questions as well as basic question types such as \category{Object Exists}. N/A indicates there were no valid compositions for a given type.}
\label{tab:treewide}
\centering
\resizebox{\linewidth}{!}{
\begin{tabular}{l rrrrr rrrr rrrr rrrr rrrr r}
                                  & \multicolumn{5}{c}{Accuracy}        & \multicolumn{4}{c}{CA} & \multicolumn{4}{c}{RWR} & \multicolumn{4}{c}{Delta} & \multicolumn{4}{c}{IC}                     &       \\ \cline{2-5} \cline{7-9} \cline{11-13} \cline{15-17} \cline{19-21}
Question Type & 
HCRN & HME & PSAC  & Most-Likely & &
HCRN & HME & PSAC  & &
HCRN & HME & PSAC  & &
HCRN & HME & PSAC  & &
HCRN & HME & PSAC  & &
Human \\ 
\hline\hline
\category{Object Exists} 
& 47.03 & 46.74 & 45.02 & 50.00 & &
N/A    & N/A   & N/A   & &
N/A    & N/A    & N/A   & &
N/A     & N/A    & N/A    & &
N/A          & N/A          & N/A          & &
92.00 \\
\category{Relation Exists}                   
& 52.14 & 51.21 & 36.44 & 50.00     &  & 
73.17  & 8.99  & N/A   & &
16.67  & N/A    & 20.22 & &
-56.50  & N/A    & N/A    & &
81.14        & N/A          & 39.89        & &
92.00 \\
\category{Interaction}                       
& 46.71 & 50.57 & 62.33 & 50.00     &  & 
62.50  & 32.66 & N/A   & &
33.31  & 23.58  & 48.63 & &
-29.19  & -9.08  & N/A    & &
74.77        & 58.54        & 32.26         & &
88.00 \\
\category{Interaction Temporal Loc.} 
& 49.53 & 50.43 & 45.20 & 50.00    &   & 
57.82  & 57.96 & 3.91  & &
47.39  & 50.46  & 46.92 & &
-10.43   & -7.51  & 43.01  & &
59.85        & 60.62        & 46.45         & &
96.00 \\
\category{Exists Temporal Loc.}      
& 47.82 & 49.69 & 53.52 & 50.00    &   & 
90.92  & 22.60 & 67.68 & &
45.44  & 1.96   & 18.69 & &
-45.49  & -20.64 & -48.99 & & 
54.24        & 75.60        & 67.14        & &
92.00 \\
\category{First/Last}                        
& 9.28  & 12.31 & 8.20  & 3.79     &   & 
N/A    & N/A   & N/A   & &
N/A    & N/A    & N/A   & &
N/A     & N/A    & N/A    & &
N/A          & N/A          & N/A          & &
88.00 \\
\category{Longest/Shortest Action}           
& 3.24  & 1.67  & 1.58  & 3.57    &    & 
N/A    & N/A   & N/A   & &
N/A    & N/A    & N/A   & &
N/A     & N/A    & N/A    & &
N/A          & N/A          & N/A          & &
76.00 \\
\category{Conjunction}                       
& 49.60 & 50.07 & 50.01 & 50.00   &    & 
71.64  & 85.26 & 85.81 & &
42.19  & 39.85  & 39.92 & &
-29.45  & -45.42 & -45.89 &  &
50.54        & 54.34        & 48.78        & &
76.00 \\
\category{Choose}                            
& 24.44 & 35.16 & 26.03 & 1.89   &     & 
51.19  & 55.24 & 46.49 & &
47.05  & 48.28  & 48.09 & &
-4.14   & -6.96  & 1.59   & &
5.75         & 0.65         & 12.18         & &
88.00 \\
\category{Equals}                            
& 50.53 & 50.08 & 49.92 & 50.00   &    & 
47.71  & 52.88 & 49.00 & &
51.67  & 47.15  & 50.36 & &
3.96    & -5.72   & 1.35   & &
28.10        & 43.35        & 39.26      &  & 
70.00 \\ \hline
Overall                           
& 21.27 & 30.47 & 21.29 & 3.31   &     & 
74.59  & 49.28 & 60.97 & &
46.22  & 25.29  & 36.68 & &
-28.37  & -23.99 & -24.28 & &
47.62 & 54.31 & 48.30 & &
84.36
\end{tabular}
}
\end{table*}

Finally, we balance the answer distribution to arrive at our final dataset. When generating AGQA's original balanced dataset, the authors used an answer smoothing algorithm to mitigate biases in the training process. Adding our sub-questions to AGQA changes the training answer distributions. To reduce the bias in the new answer distributions, we adopt the same answer smoothing algorithm. This process results in $1.62M$ unique new sub-questions across the dataset, and a total of $4.55M$ sub-questions.

\section{Metrics}\label{sec:metrics_section}
Using the sub-question types and composition rules we handcrafted, we design novel metrics that measure models' compositional accuracy, test whether models are right for the wrong reasons, and identify whether models are internally consistent. Our metrics are complementary and should be used together to guide error analysis. Formal definitions for the metrics can be found in the Supplementary.

\noindent\textbf{Compositional accuracy (\textsc{CA}):} A model reasoning compositionally should be able to answer a given parent question $q$ correctly when it answers its sub-questions correctly. We operationalize this intuition with the \textbf{CA} metric, which measures parent question accuracy across compositions where a model answers all immediate sub-questions correctly. Low CA scores for a given category indicate difficulty performing that intermediate reasoning step.

\noindent\textbf{Right for the wrong reasons (\textsc{RWR}):} Given that the sub-questions of a given question $q$ represent intermediate reasoning steps, a model reasoning compositionally should answer all sub-questions correctly if it answers $q$ correctly. Failure to do so implies the model is relying on faulty decision mechanisms to reach correct answers. The \textbf{RWR} metric aims to determine to what extent such faulty reasoning occurs. To compute this, we measure parent question accuracy across compositions where a model answers at least one sub-question incorrectly. High RWR scores for a given category imply that the model's reasoning is faulty for those intermediate steps. For granularity, we additionally compute parent question accuracies across compositions where a model answers exactly $n$ sub-questions incorrectly, where $n$ is an integer. We denote this variant \textbf{RWR-n} and present its results in the Supplementary (Tables~\ref{tab:count_treewide}, \ref{tab:count_parentchild}).

\noindent\textbf{Delta:} We derive additional insights by computing the difference between \textbf{RWR} and \textbf{CA} values. Ideally, \textbf{RWR} will be lower than \textbf{CA}, leading to negative \textbf{Delta} values. A positive \textbf{Delta} value implies incorrect reasoning since the model performs better when it errs on a sub-question.

\noindent\textbf{Internal Consistency (\textsc{IC}):} A model that reasons compositionally should produce answers that don't contradict each other, regardless of accuracy. Unlike most past work on measuring consistency~\cite{ribeiro2019red, hudson2019gqa, selvaraju2020squinting}, we can use our logical consistency rules (see Table~\ref{tab:consistencyrules} in the Supplementary) and their contrapositives to determine whether models are self-consistent without access to ground-truth answers. We note that most compositions considered for the \textbf{IC} metric have multiple logical consistency rules associated with them. To compute the \textbf{IC} metric for a given composition rule, we first measure the percentage of consistency checks a model satisfies for each of its logical consistency rules. Then we average these percentages to obtain the \textbf{IC} score for that composition. With this, we avoid overemphasizing a more common rule. \textbf{IC} scores for individual logical consistency rules can be found in the Supplementary (Table \ref{tab:ic_splits}).

\noindent\textbf{Accuracy:} To obtain a baseline understanding of model performance, we additionally compute accuracy per question type. To elevate the role of answers on the long tail of the answer distributions, we compute accuracy per ground-truth answer and then normalize across answers.

\section{Experiments}
We evaluate three state-of-the-art video question answering models on our DAGs to analyze their compositional visual reasoning capability. We start by analyzing model accuracy on leaf nodes testing basic perception. We then analyze three different groups of compositional reasoning steps: \category{Choose} and \category{Equals} questions, \category{Conjunction} questions, and the \category{Temporal Localization} categories. In these analyses, the CA metric helps determine which reasoning steps models struggle at, the RWR metric checks whether models achieve high accuracies even when failing at intermediate reasoning steps, and the IC metric determines how often models contradict themselves. We additionally cite exact values for RWR-n scores, IC values for individual consistency rules and accuracies per ground-truth answers to support analysis. Full tables for these values can be found in the Supplementary (Tables \ref{tab:count_treewide}-\ref{tab:accuracy_splits}).

\noindent\textbf{Models.} We use the three models evaluated in the AGQA paper: HME~\cite{fan2019heterogeneous}, HCRN~\cite{le2020hierarchical} and PSAC~\cite{li2019beyond}. HME fuses memory modules for visual and question features~\cite{fan2019heterogeneous}, HCRN creates a multi-layer hierarchy of a reusable module that integrates motion, question, and visual features at each layer~\cite{le2020hierarchical} and PSAC integrates visual and language features using positional self-attention and co-attention blocks~\cite{li2019beyond}. Like the AGQA paper, we also consider a model (Most-Likely) that outputs the most common answer for each question type as a baseline relying only on linguistic biases.

\noindent\textbf{Training.}
We trained models on a version of the AGQA balanced dataset that is augmented with the balanced sub-question DAGs we produced. We stop training when validation accuracy plateaues. 

\subsection{Human evaluation}

To evaluate the quality of the questions and answers our engine generates, we run a human evaluation study.
We hire annotators at a rate of $\$15$/hr in accordance with fair work standards on Amazon Mechanical Turk~\cite{whiting2019fair}. We present annotators with at least $25$ randomly sampled questions per sub-question type and adopt the human evaluation protocol presented in AGQA~\cite{GrundeMcLaughlin2021AGQA}. Annotators are asked to verify a question and answer pair by watching the video associated with them.
The majority vote of 3 annotators per question labeled $84.36\%$ of our answers as correct, implying that about $15.64\%$ of our questions contain errors (see Table~\ref{tab:treewide}). These errors originate in scene graph annotation errors and ambiguous relationships.  We describe in supplementary materials the sources of human error.
To put this number in context, GQA~\cite{hudson2019gqa}, CLEVR~\cite{johnson2017clevr} and AGQA~\cite{GrundeMcLaughlin2021AGQA}, three recent automated benchmarks, report $89.30\%$, $92.60\%$, and $86.02\%$ human accuracy, respectively.

\subsection{Performance on Leaf Nodes}
Upon inspecting model accuracy (Table \ref{tab:treewide}) on the \category{Object Exists} and \category{Relation Exists} categories, we find that each model struggles on basic perceptual questions, casting doubt on good performance on more complex categories. Model accuracy on both categories is either on par with or poorer than the Most-Likely baseline. By investigating model accuracy per-ground truth answer (see Table~\ref{tab:accuracy_splits} in the Supplementary), we find that HME is heavily biased towards ``no'' answers for \category{Relation Exists}, achieving $99.11\%$ and $3.29\%$ accuracy on ``no'' and ``yes'' answered questions respectively. PSAC is similarly biased on the \category{Object Exists} category, achieving $86.67\%$ and $3.38\%$ accuracy on ``no'' and ``yes'' answered questions. HCRN, finally, has near or below-chance performance on both categories, only achieving above $50\%$ accuracy on ``No'' answered questions of the \category{Relation Exists} category with a score of $55.84\%$.

\subsection{Performance on Choose and Equals}
Our CA, RWR and IC metrics (Table \ref{tab:parentchild}) help demonstrate not only that models struggle at the \category{Choose} and \category{Equals} categories, but that they also rely on incorrect reasoning for them. Firstly, by looking at the CA scores, we find that even when models answer all child questions correctly, they obtain around or below $50\%$ accuracy for these binary questions. Models particularly struggle at \category{Longer/Shorter Choose} compositions. HCRN, HME and PSAC, for instance, obtain $42.02\%$, $41.90\%$ and $38.51\%$ CA for \category{Longer Choose}. Furthermore, models achieve an IC score of at most $12.18\%$ for \category{Choose} compositions, providing evidence for incorrect reasoning. Models' reasoning is particularly faulty when the \category{Choose} composition requires ordering two events (Table~\ref{tab:ic_splits}), with HCRN, HME and PSAC's predictions being self-consistent only $4.92\%$, $0.54\%$ and $9.56\%$ of the time for this rule. We can reach a similar conclusion for the \category{Equals} composition. HCRN and PSAC have Delta scores of $3.96\%$ and $1.35\%$ respectively, meaning they are better at answering parent questions upon making mistakes at child questions. In contrast, HME obtains a Delta score of $-5.72\%$ (Table \ref{tab:parentchild}), indicating that errors on intermediate reasoning steps have only a small negative impact on its performance, which shouldn't occur if reasoning compositionally.

\subsection{Performance on Conjunctions}
Models' inability to reason compositionally largely persists for the logical \category{Conjunction} categories.  
While both HME and PSAC obtain high CA scores (Table \ref{tab:parentchild}) for \category{And} ($95.81\%$ and $88.31\%$) and \category{Xor} ($78.91\%$ and $84.32\%$) compositions, their success stems primarily from their performance when the parent question has ``no'' as a ground-truth answer. For the CA metric, HME and PSAC predict $41.95\%$ and $37.60\%$ of ``yes'' answered questions correctly for \category{And} compositions and only $1.41\%$ and $14.79\%$ of ``yes'' answered questions correctly for \category{Xor} compositions. Both models obtain approximately $80\%$ RWR-1 performance for \category{And} and over $80\%$ RWR-2 performance for \category{Xor} compositions (Table \ref{tab:count_parentchild}). Their performance is far above chance when making mistakes on intermediate reasoning steps, indicating that their success on ``no'' answered questions is not due to an understanding of logical conjunctions. HCRN, however, behaves differently. For \category{Xor}, it obtains a poor CA score of $52.33\%$, which is close to chance. On the other hand, HCRN appears to properly understand the \category{And} composition. It achieves a high CA score of $88.49$, answering $90.66\%$ of ``yes'' and $84.52\%$ of ``no'' answered questions correctly. Its IC score is also a high $74.04\%$ (Table \ref{tab:parentchild}), where it is internally consistent for $69.10\%$ and $78.98\%$ of consistency checks where the parent is ``yes'' and ``no'' respectively (Table \ref{tab:ic_splits}). While its RWR-1 score of $48.35$ (Table \ref{tab:count_parentchild}) casts doubt on whether HCRN has a grounded understanding of what the question asks, its high CA and IC scores nonetheless indicate that it can competently execute the \category{And} reasoning step.

 \begin{figure*}[t]
     \centering
     \includegraphics[width=0.70\linewidth]{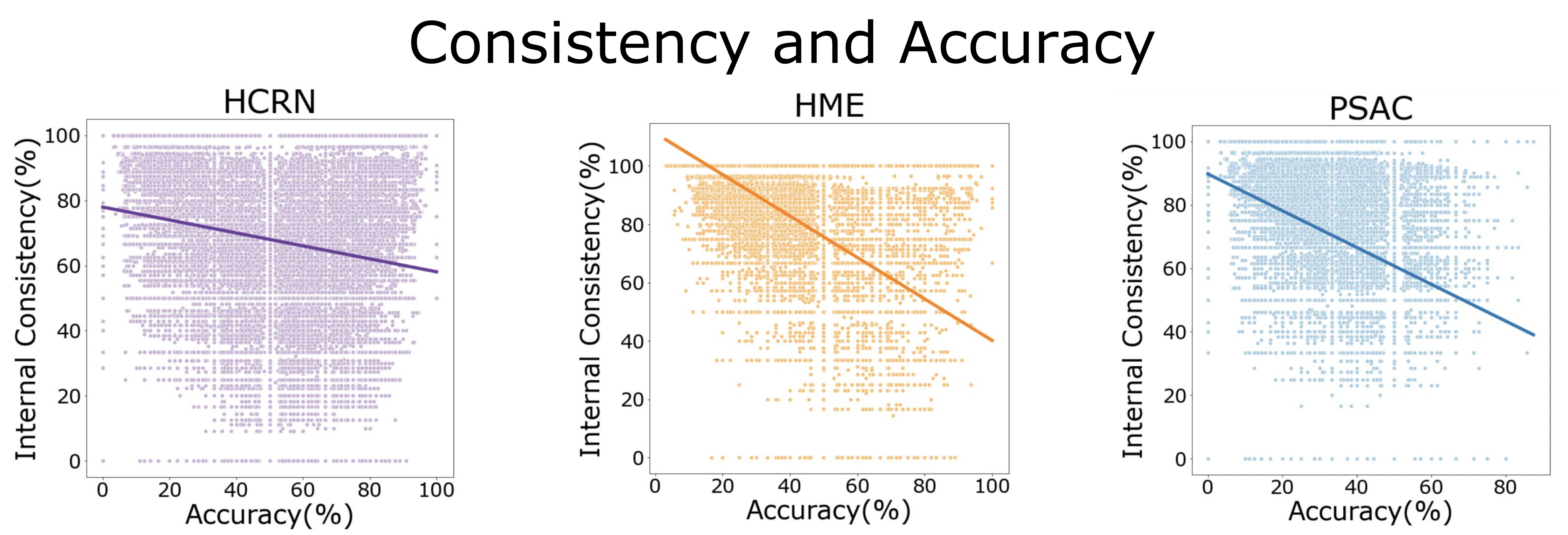}
     \caption{We measure the internal consistency of each DAG using handcrafted consistency rules. Each model has a weak negative correlation between the internal consistency of a DAG and the accuracy across all its questions. The correlation is weakest for HCRN and strongest for HME (Pearson Correlation Coefficient: $-0.206$ for HCRN, $-0.532$ for HME and $-0.424$ for PSAC).
     }
     \label{fig:consistency_accuracy}
 \end{figure*}
 
\begin{table*}[]
\caption{We calculate the compositional accuracy (\textbf{CA}), right for the wrong reasons (\textbf{RWR}), delta (\textbf{RWR-CA}) and internal consistency (\textbf{IC}) metrics with respect to composition rules for HCRN, HME and PSAC. We find that models are either unable to reason over a given composition or are right for the wrong reasons, often due to self-contradiction.
}
\label{tab:parentchild}
\centering
\resizebox{\linewidth}{!}{
\begin{tabular}{l rrrr rrrr rrrr rrr}
                 & \multicolumn{4}{c}{CA} & \multicolumn{4}{c}{RWR} & \multicolumn{4}{c}{Delta} & \multicolumn{3}{c}{IC}      \\ \cline{2-4} \cline{6-8} \cline{10-12} \cline{14-16} 
Composition Type & 
HCRN   & HME   & PSAC  & &
HCRN   & HME    & PSAC  & &
HCRN    & HME    & PSAC   & &
HCRN    & HME     & PSAC    \\ \hline\hline
\category{Interaction}      & 
58.42	& 42.09 &	92.75 &	&
40.73 &	38.85 &	49.94 & &
-17.70 &	-3.24 &	-42.82 & &	
75.42 &	61.59 &	28.32 \\
\category{First}            & 
N/A    & N/A   & N/A   & &
N/A    & N/A    & N/A   & &
N/A     & N/A    & N/A    & &
N/A     & N/A     & N/A     \\
\category{Last}             & 
N/A    & N/A   & N/A   & &
N/A    & N/A    & N/A   & &
N/A     & N/A    & N/A    & & 
N/A     & N/A     & N/A     \\
\category{Equals}           & 
47.71 &	52.88 &	49.00 & &
51.67 &	47.15 &	50.36 & &
3.96 &	-5.72 &	1.35 & &
28.10 &	43.35 &	39.26  \\
\category{And}              & 
88.49  & 95.81 & 88.31 & &
34.86  & 40.79  & 42.46 & &
-53.63  & -55.03 & -45.85 & &
74.04 & 64.12 & 48.05 \\
\category{Xor}              & 
52.33  & 78.91 & 84.32 & &
49.21  & 38.76  & 36.98 & &
-3.12   & -40.15 & -47.34 & &
27.04 & 44.56 & 49.51 \\
\category{Choose}           & 
52.42 &	57.02 &	47.64 & &
47.42 &	48.58 &	48.32 & &
-5.00 & -8.45 &	0.68 & & 
5.75 &	0.65 &	12.18 \\
\category{Longer Choose}    & 
42.02 &	41.90 &	38.51 & &
38.68 &	41.04 &	41.00 & &
-3.34 &	-0.87 &	2.49 & & 
N/A     & N/A     & N/A     \\
\category{Shorter Choose}   & 
40.28 &	50.88 &	38.86 & &
36.83 &	41.87 &	41.44 & &
-3.45 &	-9.01 &	2.58 & &
N/A     & N/A     & N/A     \\
\category{After}            & 
78.10 &	30.88 &	57.34 & &
48.02 & 22.45 &	30.00 & &
-30.08 & -8.43 & -27.34 & &
47.41 &	69.55 &	56.32 \\
\category{Before}           & 
78.49 &	31.95 &	58.48 &	&
51.93 &	21.93 &	28.73 & &
-26.57 & -10.02 & -29.75 & &
43.96 &	70.43 &	57.17 \\
\category{While}            & 
89.36 &	24.36 &	64.53 & &
44.40 &	9.33 &	23.88 & &
-44.96 & -15.03 & -40.66 & &
51.30 &	66.27 &	62.39 \\
\category{Between}          & 
84.80  & 94.54 & 89.38 & &
17.37  & 5.85   & 12.25 & &
-67.43  & -88.69 & -77.12 & &
75.56 & 68.24 & 81.54 \\ \hline
Overall         & 
69.70 &	51.90 &	62.29 & &
45.84 &	27.98 &	37.82 & &
-23.87 & -23.92 & -24.47 & &
47.62 &	54.31 &	48.30
\end{tabular}
}
\end{table*}

\subsection{Performance on Temporal Reasoning}
We finally analyze model performances on the \category{Temporal Localization} categories, starting with the \category{Exists Temporal Localization} question type. We split analysis by the temporal localization composition types: \category{After}, \category{Before}, \category{While} or \category{Between}. We first find that HME fails on \category{After}, \category{Before} and \category{While} compositions, obtaining poor CA scores of $30.88\%$, $31.95\%$ and $24.36\%$ respectively (Table \ref{tab:parentchild}). While PSAC and particularly HCRN obtain higher CA scores on these compositions, their success is likely due to faulty reasoning. Both models obtain IC scores less than $50\%$ when answering ``yes'' to the parent question (Table \ref{tab:ic_splits}), contradicting themselves over half the time in one common setting. HCRN's above chance RWR-1 scores of $61.24\%$, $65.16\%$, $66.01\%$ for these compositions (Table \ref{tab:count_parentchild}) further indicate incorrect reasoning. Model performances on \category{Between} compositions, however, are reminiscent of those on \category{And} compositions. While HME obtains a high CA score of $94.54\%$, it achieving an IC score of $37.72\%$ when the parent is ``yes'' (Table \ref{tab:ic_splits}) and a high RWR-1 score of $77.83\%$ (Table \ref{tab:count_parentchild}) indicates that this success is due to incorrect reasoning. Meanwhile, HCRN and PSAC achieve high CA scores, do not have RWR values far above chance (Tables \ref{tab:parentchild}, \ref{tab:count_parentchild}) and obtain high IC scores of $75.56\%$ and $81.54\%$ respectively. These models can successfully execute the \category{Between} reasoning step even if their understandings of the underlying \category{Before} and \category{After} compositions are suspect. \category{Interaction Temporal Localization}, on the other hand, additionally involves an \category{Interaction} composition and requires the model to temporally reason about two different relationships or actions. PSAC, given its $3.91\%$ CA score, is incapable of performing this task. HCRN and HME, on the other hand, likely rely on spurious correlations even when they are correct. For instance, while HCRN and HME obtain CA scores of $57.82\%$ and $57.96\%$ respectively (Table \ref{tab:parentchild}), they also obtain RWR-2 scores of $55.34\%$ and $93.92\%$ (Table \ref{tab:count_parentchild}), meaning that their performance does not depend on whether they are accurate for intermediate reasoning steps. Models' overall poor performance on \category{Interaction Temporal Localization} is similar to the performance on \category{Choose} and \category{Equals} questions, both of which also require reasoning over two distinct components.

\subsection{Correlation between consistency and accuracy}
We test whether our IC metric is predictive of model accuracy, as this can aid users at inference time. Specifically, we measure whether IC is correlated with accuracy. To do this, we compute internal consistency on DAGs by measuring the percentage of correct logical consistency checks across all compositions in a DAG and compare against accuracy on the entire DAG. Figure~\ref{fig:consistency_accuracy} shows that internal consistency has a weak negative correlation with accuracy, with HCRN, HME and PSAC having correlation coefficients of $-0.206$, $-0.532$ and $-0.424$ respectively. HME's stronger correlation can be explained by its consistent bias towards ``no'' answers (see Table \ref{tab:accuracy_splits} in the Supplementary), which are less frequent in our DAGs as our consistency checks can only propagate ``yes'' answers. As such, while HME is highly consistent, it is also frequently incorrect, which causes inaccuracies to propagate throughout hierarchies. PSAC shares HME's bias towards ``no'' answers for some question categories, causing it to obtain a more negative correlation than HCRN. Finally, HCRN is a less biased model that is often right for the wrong reasons. Thus, being internally consistent does not imply being accurate for HCRN.

\section{Discussion}
In conclusion, we developed a question decomposition engine and generated the dataset AGQA-Decomp hoping to facilitate the analysis of video question answering models beyond average accuracy. 
Our work is a continuation of a shift in machine learning away from standard accuracy metrics towards more metamorphic evaluation~\cite{gardner2020evaluating,bitton2021automatic,li2020linguistically}. Our results are bleak: models frequently contradict themselves and are often right for the wrong reasons. 

\paragraph{Acknowledgements.} This work was partially supported by the Brown Institute for Media Innovation. We also thank Jerry Hong, Zixian Ma and Helena Vasconcelos for their valuable insights.

{\small
\bibliographystyle{ieee_fullname}
\bibliography{egbib}

\begin{thebibliography}{10}\itemsep=-1pt

\bibitem{andreas2016neural}
Jacob Andreas, Marcus Rohrbach, Trevor Darrell, and Dan Klein.
\newblock Neural module networks.
\newblock In {\em Proceedings of the IEEE conference on computer vision and
  pattern recognition}, pages 39--48, 2016.

\bibitem{birhane2021multimodal}
Abeba Birhane, Vinay~Uday Prabhu, and Emmanuel Kahembwe.
\newblock Multimodal datasets: misogyny, pornography, and malignant
  stereotypes.
\newblock {\em arXiv preprint arXiv:2110.01963}, 2021.

\bibitem{bitton2021automatic}
Yonatan Bitton, Gabriel Stanovsky, Roy Schwartz, and Michael Elhadad.
\newblock Automatic generation of contrast sets from scene graphs: Probing the
  compositional consistency of gqa.
\newblock {\em arXiv preprint arXiv:2103.09591}, 2021.

\bibitem{buolamwini2018gender}
Joy Buolamwini and Timnit Gebru.
\newblock Gender shades: Intersectional accuracy disparities in commercial
  gender classification.
\newblock In {\em Conference on fairness, accountability and transparency},
  pages 77--91. PMLR, 2018.

\bibitem{cao2018visual}
Qingxing Cao, Xiaodan Liang, Bailing Li, Guanbin Li, and Liang Lin.
\newblock Visual question reasoning on general dependency tree.
\newblock In {\em Proceedings of the IEEE Conference on Computer Vision and
  Pattern Recognition}, pages 7249--7257, 2018.

\bibitem{chen2021meta}
Wenhu Chen, Zhe Gan, Linjie Li, Yu Cheng, William Wang, and Jingjing Liu.
\newblock Meta module network for compositional visual reasoning.
\newblock In {\em Proceedings of the IEEE/CVF Winter Conference on Applications
  of Computer Vision}, pages 655--664, 2021.

\bibitem{chomsky2002syntactic}
Noam Chomsky.
\newblock {\em Syntactic structures}.
\newblock Walter de Gruyter, 2002.

\bibitem{cresswell1973logics}
MJ Cresswell.
\newblock Logics and languages.
\newblock 1973.

\bibitem{dalvi2021explaining}
Bhavana Dalvi, Peter Jansen, Oyvind Tafjord, Zhengnan Xie, Hannah Smith,
  Leighanna Pipatanangkura, and Peter Clark.
\newblock Explaining answers with entailment trees.
\newblock {\em arXiv preprint arXiv:2104.08661}, 2021.

\bibitem{fan2019heterogeneous}
Chenyou Fan, Xiaofan Zhang, Shu Zhang, Wensheng Wang, Chi Zhang, and Heng
  Huang.
\newblock Heterogeneous memory enhanced multimodal attention model for video
  question answering.
\newblock In {\em Proceedings of the IEEE/CVF conference on computer vision and
  pattern recognition}, pages 1999--2007, 2019.

\bibitem{gardner2020evaluating}
Matt Gardner, Yoav Artzi, Victoria Basmova, Jonathan Berant, Ben Bogin, Sihao
  Chen, Pradeep Dasigi, Dheeru Dua, Yanai Elazar, Ananth Gottumukkala, et~al.
\newblock Evaluating models' local decision boundaries via contrast sets.
\newblock {\em arXiv preprint arXiv:2004.02709}, 2020.

\bibitem{Girdhar2020CATER:}
Rohit Girdhar and Deva Ramanan.
\newblock Cater: A diagnostic dataset for compositional actions \& temporal
  reasoning.
\newblock In {\em International Conference on Learning Representations}, 2020.

\bibitem{gokhale2020vqa}
Tejas Gokhale, Pratyay Banerjee, Chitta Baral, and Yezhou Yang.
\newblock Vqa-lol: Visual question answering under the lens of logic.
\newblock In {\em European conference on computer vision}, pages 379--396.
  Springer, 2020.

\bibitem{GrundeMcLaughlin2021AGQA}
Madeleine Grunde-McLaughlin, Ranjay Krishna, and Maneesh Agrawala.
\newblock Agqa: A benchmark for compositional spatio-temporal reasoning.
\newblock In {\em Proceedings of the IEEE/CVF Conference on Computer Vision and
  Pattern Recognition}, 2021.

\bibitem{hendricks2016generating}
Lisa~Anne Hendricks, Zeynep Akata, Marcus Rohrbach, Jeff Donahue, Bernt
  Schiele, and Trevor Darrell.
\newblock Generating visual explanations.
\newblock In {\em European conference on computer vision}, pages 3--19.
  Springer, 2016.

\bibitem{hendricks2018women}
Lisa~Anne Hendricks, Kaylee Burns, Kate Saenko, Trevor Darrell, and Anna
  Rohrbach.
\newblock Women also snowboard: Overcoming bias in captioning models.
\newblock In {\em Proceedings of the European Conference on Computer Vision
  (ECCV)}, pages 771--787, 2018.

\bibitem{hendricks2018grounding}
Lisa~Anne Hendricks, Ronghang Hu, Trevor Darrell, and Zeynep Akata.
\newblock Grounding visual explanations.
\newblock In {\em Proceedings of the European Conference on Computer Vision
  (ECCV)}, pages 264--279, 2018.

\bibitem{hu2017learning}
Ronghang Hu, Jacob Andreas, Marcus Rohrbach, Trevor Darrell, and Kate Saenko.
\newblock Learning to reason: End-to-end module networks for visual question
  answering.
\newblock In {\em Proceedings of the IEEE International Conference on Computer
  Vision}, pages 804--813, 2017.

\bibitem{hudson2019gqa}
Drew~A Hudson and Christopher~D Manning.
\newblock Gqa: A new dataset for real-world visual reasoning and compositional
  question answering.
\newblock In {\em Proceedings of the IEEE/CVF conference on computer vision and
  pattern recognition}, pages 6700--6709, 2019.

\bibitem{jang2017tgif}
Yunseok Jang, Yale Song, Youngjae Yu, Youngjin Kim, and Gunhee Kim.
\newblock Tgif-qa: Toward spatio-temporal reasoning in visual question
  answering.
\newblock In {\em Proceedings of the IEEE conference on computer vision and
  pattern recognition}, pages 2758--2766, 2017.

\bibitem{ji2020action}
Jingwei Ji, Ranjay Krishna, Li Fei-Fei, and Juan~Carlos Niebles.
\newblock Action genome: Actions as compositions of spatio-temporal scene
  graphs.
\newblock In {\em Proceedings of the IEEE/CVF Conference on Computer Vision and
  Pattern Recognition}, pages 10236--10247, 2020.

\bibitem{johnson2017clevr}
Justin Johnson, Bharath Hariharan, Laurens Van Der~Maaten, Li Fei-Fei, C
  Lawrence~Zitnick, and Ross Girshick.
\newblock Clevr: A diagnostic dataset for compositional language and elementary
  visual reasoning.
\newblock In {\em Proceedings of the IEEE conference on computer vision and
  pattern recognition}, pages 2901--2910, 2017.

\bibitem{kim2017deepstory}
Kyung-Min Kim, Min-Oh Heo, Seong-Ho Choi, and Byoung-Tak Zhang.
\newblock Deepstory: video story qa by deep embedded memory networks.
\newblock In {\em Proceedings of the 26th International Joint Conference on
  Artificial Intelligence}, pages 2016--2022, 2017.

\bibitem{krishna2017visual}
Ranjay Krishna, Yuke Zhu, Oliver Groth, Justin Johnson, Kenji Hata, Joshua
  Kravitz, Stephanie Chen, Yannis Kalantidis, Li-Jia Li, David~A Shamma, et~al.
\newblock Visual genome: Connecting language and vision using crowdsourced
  dense image annotations.
\newblock {\em International journal of computer vision}, 123(1):32--73, 2017.

\bibitem{kurby2008segmentation}
Christopher~A Kurby and Jeffrey~M Zacks.
\newblock Segmentation in the perception and memory of events.
\newblock {\em Trends in cognitive sciences}, 12(2):72--79, 2008.

\bibitem{lake2018generalization}
Brenden Lake and Marco Baroni.
\newblock Generalization without systematicity: On the compositional skills of
  sequence-to-sequence recurrent networks.
\newblock In {\em International Conference on Machine Learning}, pages
  2873--2882, 2018.

\bibitem{le2020hierarchical}
Thao~Minh Le, Vuong Le, Svetha Venkatesh, and Truyen Tran.
\newblock Hierarchical conditional relation networks for video question
  answering.
\newblock In {\em Proceedings of the IEEE/CVF conference on computer vision and
  pattern recognition}, pages 9972--9981, 2020.

\bibitem{lei2018tvqa}
Jie Lei, Licheng Yu, Mohit Bansal, and Tamara~L Berg.
\newblock Tvqa: Localized, compositional video question answering.
\newblock {\em arXiv preprint arXiv:1809.01696}, 2018.

\bibitem{li2020linguistically}
Chuanrong Li, Lin Shengshuo, Leo~Z Liu, Xinyi Wu, Xuhui Zhou, and Shane
  Steinert-Threlkeld.
\newblock Linguistically-informed transformations (lit): A method for
  automatically generating contrast sets.
\newblock {\em arXiv preprint arXiv:2010.08580}, 2020.

\bibitem{li2019beyond}
Xiangpeng Li, Jingkuan Song, Lianli Gao, Xianglong Liu, Wenbing Huang, Xiangnan
  He, and Chuang Gan.
\newblock Beyond rnns: Positional self-attention with co-attention for video
  question answering.
\newblock In {\em Proceedings of the AAAI Conference on Artificial
  Intelligence}, volume~33, pages 8658--8665, 2019.

\bibitem{lillo2014discriminative}
Ivan Lillo, Alvaro Soto, and Juan Carlos~Niebles.
\newblock Discriminative hierarchical modeling of spatio-temporally composable
  human activities.
\newblock In {\em Proceedings of the IEEE conference on computer vision and
  pattern recognition}, pages 812--819, 2014.

\bibitem{min2019multi}
Sewon Min, Victor Zhong, Luke Zettlemoyer, and Hannaneh Hajishirzi.
\newblock Multi-hop reading comprehension through question decomposition and
  rescoring.
\newblock {\em arXiv preprint arXiv:1906.02916}, 2019.

\bibitem{montague1970universal}
Richard Montague et~al.
\newblock Universal grammar.
\newblock {\em 1974}, pages 222--46, 1970.

\bibitem{mun2017marioqa}
Jonghwan Mun, Paul Hongsuck~Seo, Ilchae Jung, and Bohyung Han.
\newblock Marioqa: Answering questions by watching gameplay videos.
\newblock In {\em Proceedings of the IEEE International Conference on Computer
  Vision}, pages 2867--2875, 2017.

\bibitem{perez2020unsupervised}
Ethan Perez, Patrick Lewis, Wen-tau Yih, Kyunghyun Cho, and Douwe Kiela.
\newblock Unsupervised question decomposition for question answering.
\newblock {\em arXiv preprint arXiv:2002.09758}, 2020.

\bibitem{ray2019sunny}
Arijit Ray, Karan Sikka, Ajay Divakaran, Stefan Lee, and Giedrius Burachas.
\newblock Sunny and dark outside?! improving answer consistency in vqa through
  entailed question generation.
\newblock {\em arXiv preprint arXiv:1909.04696}, 2019.

\bibitem{reynolds2007computational}
Jeremy~R Reynolds, Jeffrey~M Zacks, and Todd~S Braver.
\newblock A computational model of event segmentation from perceptual
  prediction.
\newblock {\em Cognitive science}, 31(4):613--643, 2007.

\bibitem{ribeiro2019red}
Marco~Tulio Ribeiro, Carlos Guestrin, and Sameer Singh.
\newblock Are red roses red? evaluating consistency of question-answering
  models.
\newblock In {\em Proceedings of the 57th Annual Meeting of the Association for
  Computational Linguistics}, pages 6174--6184, 2019.

\bibitem{ribeiro2016should}
Marco~Tulio Ribeiro, Sameer Singh, and Carlos Guestrin.
\newblock " why should i trust you?" explaining the predictions of any
  classifier.
\newblock In {\em Proceedings of the 22nd ACM SIGKDD international conference
  on knowledge discovery and data mining}, pages 1135--1144, 2016.

\bibitem{selvaraju2017grad}
Ramprasaath~R Selvaraju, Michael Cogswell, Abhishek Das, Ramakrishna Vedantam,
  Devi Parikh, and Dhruv Batra.
\newblock Grad-cam: Visual explanations from deep networks via gradient-based
  localization.
\newblock In {\em Proceedings of the IEEE international conference on computer
  vision}, pages 618--626, 2017.

\bibitem{selvaraju2020squinting}
Ramprasaath~R Selvaraju, Purva Tendulkar, Devi Parikh, Eric Horvitz,
  Marco~Tulio Ribeiro, Besmira Nushi, and Ece Kamar.
\newblock Squinting at vqa models: Introspecting vqa models with sub-questions.
\newblock In {\em Proceedings of the IEEE/CVF Conference on Computer Vision and
  Pattern Recognition}, pages 10003--10011, 2020.

\bibitem{shah2019cycle}
Meet Shah, Xinlei Chen, Marcus Rohrbach, and Devi Parikh.
\newblock Cycle-consistency for robust visual question answering.
\newblock In {\em Proceedings of the IEEE/CVF Conference on Computer Vision and
  Pattern Recognition}, pages 6649--6658, 2019.

\bibitem{sigurdsson2016hollywood}
Gunnar~A Sigurdsson, G{\"u}l Varol, Xiaolong Wang, Ali Farhadi, Ivan Laptev,
  and Abhinav Gupta.
\newblock Hollywood in homes: Crowdsourcing data collection for activity
  understanding.
\newblock In {\em European Conference on Computer Vision}, pages 510--526.
  Springer, 2016.

\bibitem{speer2007human}
Nicole~K Speer, Jeffrey~M Zacks, and Jeremy~R Reynolds.
\newblock Human brain activity time-locked to narrative event boundaries.
\newblock {\em Psychological Science}, 18(5):449--455, 2007.

\bibitem{tapaswi2016movieqa}
Makarand Tapaswi, Yukun Zhu, Rainer Stiefelhagen, Antonio Torralba, Raquel
  Urtasun, and Sanja Fidler.
\newblock Movieqa: Understanding stories in movies through question-answering.
\newblock In {\em Proceedings of the IEEE conference on computer vision and
  pattern recognition}, pages 4631--4640, 2016.

\bibitem{vatashsky2020vqa}
Ben-Zion Vatashsky and Shimon Ullman.
\newblock Vqa with no questions-answers training.
\newblock In {\em Proceedings of the IEEE/CVF Conference on Computer Vision and
  Pattern Recognition}, pages 10376--10386, 2020.

\bibitem{wang2020revise}
Angelina Wang, Arvind Narayanan, and Olga Russakovsky.
\newblock Revise: A tool for measuring and mitigating bias in visual datasets.
\newblock In {\em European Conference on Computer Vision}, pages 733--751.
  Springer, 2020.

\bibitem{whiting2019fair}
Mark~E Whiting, Grant Hugh, and Michael~S Bernstein.
\newblock Fair work: Crowd work minimum wage with one line of code.
\newblock In {\em Proceedings of the AAAI Conference on Human Computation and
  Crowdsourcing}, volume~7, pages 197--206, 2019.

\bibitem{wolfson2020break}
Tomer Wolfson, Mor Geva, Ankit Gupta, Matt Gardner, Yoav Goldberg, Daniel
  Deutch, and Jonathan Berant.
\newblock Break it down: A question understanding benchmark.
\newblock {\em Transactions of the Association for Computational Linguistics},
  8:183--198, 2020.

\bibitem{wu2019errudite}
Tongshuang Wu, Marco~Tulio Ribeiro, Jeffrey Heer, and Daniel~S Weld.
\newblock Errudite: Scalable, reproducible, and testable error analysis.
\newblock In {\em Proceedings of the 57th Annual Meeting of the Association for
  Computational Linguistics}, pages 747--763, 2019.

\bibitem{wu2021polyjuice}
Tongshuang Wu, Marco~Tulio Ribeiro, Jeffrey Heer, and Daniel~S Weld.
\newblock Polyjuice: Generating counterfactuals for explaining, evaluating, and
  improving models.
\newblock In {\em Proceedings of the 59th Annual Meeting of the Association for
  Computational Linguistics}, 2021.

\bibitem{yang2018hotpotqa}
Zhilin Yang, Peng Qi, Saizheng Zhang, Yoshua Bengio, William~W Cohen, Ruslan
  Salakhutdinov, and Christopher~D Manning.
\newblock Hotpotqa: A dataset for diverse, explainable multi-hop question
  answering.
\newblock {\em arXiv preprint arXiv:1809.09600}, 2018.

\bibitem{yi2019clevrer}
Kexin Yi, Chuang Gan, Yunzhu Li, Pushmeet Kohli, Jiajun Wu, Antonio Torralba,
  and Joshua~B Tenenbaum.
\newblock Clevrer: Collision events for video representation and reasoning.
\newblock {\em arXiv preprint arXiv:1910.01442}, 2019.

\bibitem{Yi*2020CLEVRER:}
Kexin Yi*, Chuang Gan*, Yunzhu Li, Pushmeet Kohli, Jiajun Wu, Antonio Torralba,
  and Joshua~B. Tenenbaum.
\newblock Clevrer: Collision events for video representation and reasoning.
\newblock In {\em International Conference on Learning Representations}, 2020.

\bibitem{yu2019activitynet}
Zhou Yu, Dejing Xu, Jun Yu, Ting Yu, Zhou Zhao, Yueting Zhuang, and Dacheng
  Tao.
\newblock Activitynet-qa: A dataset for understanding complex web videos via
  question answering.
\newblock In {\em Proceedings of the AAAI Conference on Artificial
  Intelligence}, volume~33, pages 9127--9134, 2019.

\bibitem{yuan2021perception}
Yuanyuan Yuan, Shuai Wang, Mingyue Jiang, and Tsong~Yueh Chen.
\newblock Perception matters: Detecting perception failures of vqa models using
  metamorphic testing.
\newblock In {\em Proceedings of the IEEE/CVF Conference on Computer Vision and
  Pattern Recognition}, pages 16908--16917, 2021.

\bibitem{zadeh2019social}
Amir Zadeh, Michael Chan, Paul~Pu Liang, Edmund Tong, and Louis-Philippe
  Morency.
\newblock Social-iq: A question answering benchmark for artificial social
  intelligence.
\newblock In {\em Proceedings of the IEEE Conference on Computer Vision and
  Pattern Recognition}, pages 8807--8817, 2019.

\bibitem{zhao2021understanding}
Dora Zhao, Angelina Wang, and Olga Russakovsky.
\newblock Understanding and evaluating racial biases in image captioning.
\newblock In {\em Proceedings of the IEEE/CVF International Conference on
  Computer Vision}, pages 14830--14840, 2021.

\bibitem{zhao-etal-2017-men}
Jieyu Zhao, Tianlu Wang, Mark Yatskar, Vicente Ordonez, and Kai-Wei Chang.
\newblock Men also like shopping: Reducing gender bias amplification using
  corpus-level constraints.
\newblock In {\em Proceedings of the 2017 Conference on Empirical Methods in
  Natural Language Processing}, pages 2979--2989, Copenhagen, Denmark, Sept.
  2017. Association for Computational Linguistics.

\end{thebibliography}
}

\setcounter{section}{6}
\setcounter{figure}{3}
\setcounter{table}{4}
\section{Supplementary}

The supplementary sections provide more detail on the methods and experiments described in our paper. First, we explain in more detail our process for decomposing questions into hierarchies, including two human studies. We provide equations for our metrics, then describe our training process for the experiments, explore how AGQA-Decomp performs as data augmentation for AGQA, add results for the Most Likely baseline, and present example error modes we found through qualitative analysis of hierarchies. Finally, we discuss directions for future work.

\subsection{Dataset}

In this section, we provide additional details for the process of question decomposition. We first describe the process of generating individual sub-question hierarchies from AGQA programs. We then explain how we obtain answers for these questions, first detailing the general case and then describing the edge case of Object Exists questions. We finally discuss the limitations and potential societal impact of our hierarchies and report human performance measured through AMT studies.

\SetKwInput{KwInput}{Input}                
\SetKwInput{KwOutput}{Output}              
\SetKwProg{main}{main}{ is}{end}

\SetKwProg{Def}{def}{:}{}

\begin{algorithm}[b]
    \DontPrintSemicolon
    \KwInput{$p$: Question program}
    \KwOutput{Question decomposition hierarchy}
    \SetAlgoLined
    \Def{main(p)}{
        $V$ = empty set for vertices \;
        $E$ = empty set for edges\;
        buildDAG($p$)\;
    }
    \;
    
    \Def{buildDAG(p)}{
        $subprograms$ = inner functions of $p$\;
        \If{no subprograms} {
            $s$ = $p$'s natural language question equivalent\;
            Add $s$ to $V$ \;
            $indirect$ = program phrase replacing $p$\;
            return $s$, $indirect$\;
        }\;
        $S_q$ = empty set for subquestions\;
        \For{subprogram in subprograms}{
            $s$, $indirect$ = buildDAG($subprogram$)\;
            Add $s$ to $S_q$\;
            $p$ = $p$ replacing $subprogram$ with $indirect$\;
        }
        \;
        $q$ = $p$'s natural language question\;
        Add $q$ to $V$ \;
        
        \For{$s$ in $S_q$}{
            Add ($q$, $s$, composition) to $E$
        }

        return question, reference

    }

\caption{Question hierarchy generation}
\label{alg:question_creation}
\end{algorithm}

\begin{figure*}[t]
     \centering
     \includegraphics[width=\linewidth]{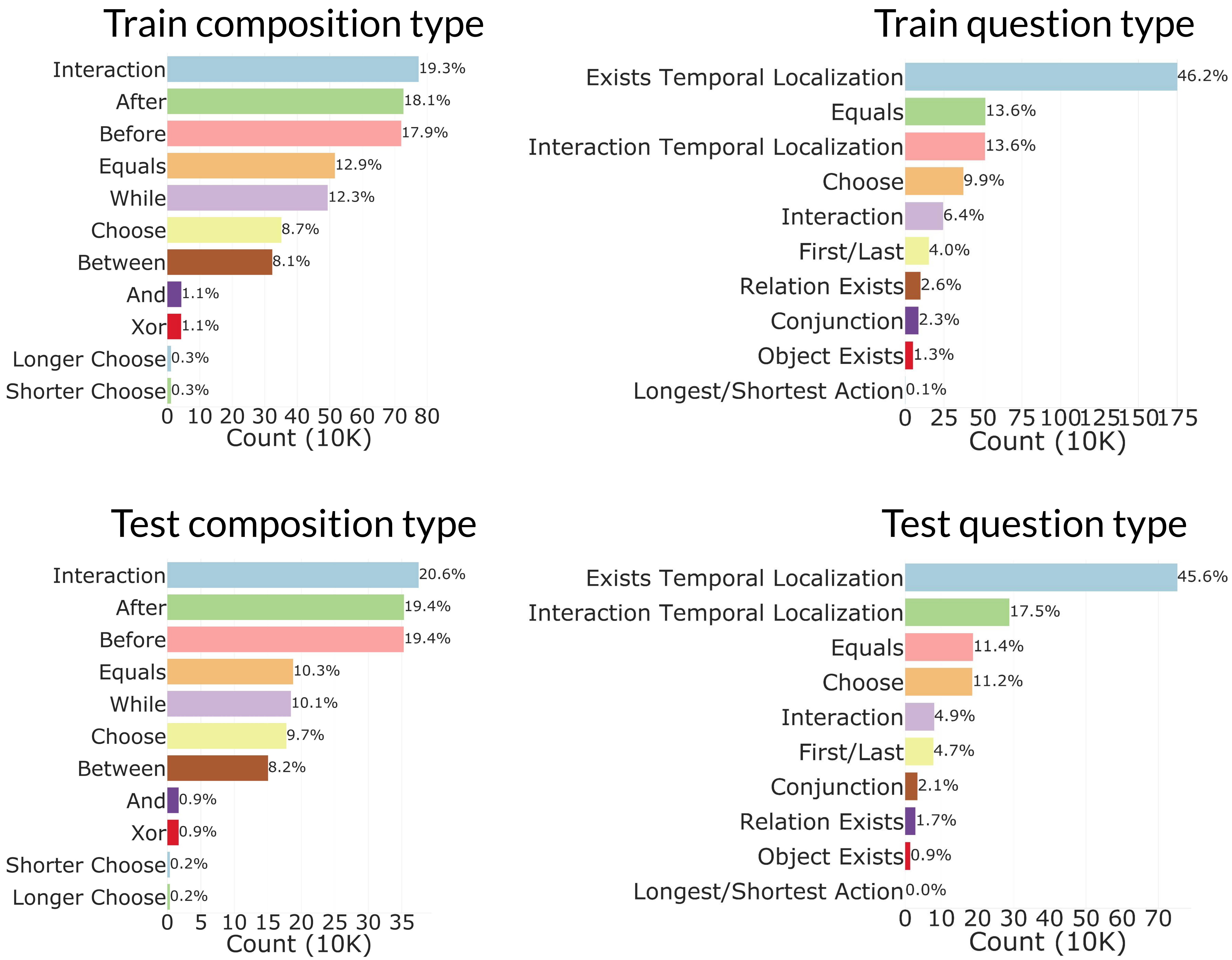}
     \caption{\textbf{Left:} A bar chart displaying the distribution of composition rule types on the test set. Interaction and After composition rules are the most common. \textbf{Right:} A bar chart displaying the distribution of question types on the test set. Exists temporal localization questions dominate the test set.}
     \label{fig:stats}
 \end{figure*}
 
 \begin{table*}[]
\caption{We handcraft logical consistency rules that check whether a model is consistent when answering questions in a DAG. The rules are implications, i.e if \textit{q} has answer \textit{a} then \textit{s1} should be \textit{b}.}
\label{tab:consistencyrules}
\centering
\resizebox{\linewidth}{!}{
\begin{tabular}{lll}

\textbf{Composition}           & \textbf{Consistency Rules}                                                                                                                                                                                                                                                        & \textbf{Example}                                                                                                                                                                                                                                                                \\ \hline
\category{Interaction}           & \begin{tabular}[c]{@{}l@{}}If an interaction `{[}\object{person}{]} {[}\relationship{relation}{]} {[}\object{object}{]}' is ‘Yes’\\ its direct sub-questions `{[}\object{person}{]}' exist, \\ `{[}\relationship{relationship}{]}' exist and \\ 
{[}\object{object}{]}' should be ‘Yes’\end{tabular}                                                                                                                                       & \begin{tabular}[c]{@{}l@{}}\emph{q }: Is a \object{person} \relationship{holding} a \object{dish}? -- Yes \\ \emph{s1}: Does a \object{person} exist? -- Yes\\ \emph{s2}: Is a person \relationship{holding} something? -- Yes\\ \emph{s3}: Does a \object{dish} exist? -- Yes\end{tabular}                                                                        \\ \hline
\category{Temporal localization} & \begin{tabular}[c]{@{}l@{}}If ‘{[}exists question{]} {[}\temporal{temporal localization}{]} {[}condition{]}’ is ‘Yes’ \\ then `{[}exists question{]}' is ‘Yes’  \\ and `{[}condition{]}' is ‘Yes’\end{tabular}                                                                & \begin{tabular}[c]{@{}l@{}}\emph{q }: Does a \object{person} exist \temporal{after} \action{smiling at something?} -- Yes \\ \emph{s1}: Does a \object{person} exist? -- Yes\\ \emph{s2}: Is a \object{person} \action{smiling at something}? -- Yes\end{tabular}                                                                                 \\ \hline
\category{And} & \begin{tabular}[c]{@{}l@{}}If ‘{[}\action{action1}{]} \conj{and} {[}\action{action2}{]}’ is ‘Yes’ \\ then `{[}\action{action1}{]}' should be ‘Yes’ \\ and `{[}\action{action2}{]}' should be ‘Yes’\end{tabular}                                                                                               & \begin{tabular}[c]{@{}l@{}}\emph{q }: Is the \object{person} \action{holding a cup} \conj{and} \action{touching a dish}? -- Yes \\ \emph{s1}: Is the \object{person} \action{holding a cup}? -- Yes   AND\\ \emph{s2}: Is the \object{person} \action{touching a dish}? -- Yes\end{tabular}                                             \\\cline{2-3}
& \begin{tabular}[c]{@{}l@{}}If ‘{[}\action{action1}{]} \conj{and} {[}\action{action2}{]}’ is ‘No’ then\\     either {[}\action{action1}{]} should be ‘No’ \\     or {[}\action{action2}{]} should be ‘No’\end{tabular}                                       & \begin{tabular}[c]{@{}l@{}}\emph{q} : Is the \object{person} \action{touching a bottle} \conj{and} \action{opening a window}? -- No \\ \emph{s1}: Is the \object{person} \action{touching a bottle}? -- No    \textbf{OR} \\ \emph{s2}: Is the \object{person} \action{opening a window}? -- No\end{tabular}                       \\ \hline
\category{Xor}                   &\begin{tabular}[c]{@{}l@{}}If ‘{[}\action{action1}{]} \conj{but not} {[}\action{action2}{]}’ is ‘Yes’ \\ then `{[}\action{action1}{]}' should be ‘Yes’  \\ and `{[}\action{action2}{]}' should be ‘No’\end{tabular}                                                                                           & \begin{tabular}[c]{@{}l@{}}\emph{q }: Is the \object{person} \action{smiling at something} \conj{but not} \action{walking through a doorway}? -- Yes \\ \emph{s1}: Is the \object{person} \action{smiling at something}? -- Yes \\ \emph{s2}: Is the \object{person} \action{walking through a doorway}? -- No\end{tabular}                                       \\\cline{2-3}
& \begin{tabular}[c]{@{}l@{}}If ‘{[}\action{action1}{]} \conj{but not} {[}\action{action2}{]}’ is ‘No’ then \\ either {[}\action{action1}{]} should be ‘No’ \\ or {[}\action{action2}{]} should be ‘Yes’\end{tabular}                                             & \begin{tabular}[c]{@{}l@{}}\emph{q }:Is the \object{person} \action{throwing a cup} \conj{but not} \action{leaning on the doorway}? -- No \\\emph{s1}: Is the \object{person} \action{throwing a cup}? -- No \textbf{OR}\\ \emph{s2}: Is the \object{person} \action{leaning on a doorway}? -- Yes\end{tabular}           \\ \hline
\category{Equals}        & \begin{tabular}[c]{@{}l@{}}If ‘{[}\object{object}{]} equals {[}\object{indirect object}{]}’ is ‘Yes’ \\ then `{[}\object{indirect object}{]}' should be `{[}\object{object}{]}'  \\ and `{[}\object{object}{]}' exists is ‘Yes’\end{tabular}                                                                  & \begin{tabular}[c]{@{}l@{}}\emph{q }: Is a \object{doorway} the \object{first object they are holding}? -- Yes \\ \emph{s1}: Which is the \object{first object they are holding}? -- \object{doorway} \\ \emph{s2}: Does a \object{doorway} exist? -- Yes\end{tabular}                        \\\cline{2-3}
& \begin{tabular}[c]{@{}l@{}}If ‘{[}\object{object}{]} equals {[}\object{indirect object}{]}’ is ‘No’ then \\ {[}\object{indirect object}{]} should not be {[}\object{object}{]}\end{tabular}                                                              & \begin{tabular}[c]{@{}l@{}}\emph{q }: Is the \object{book} the \object{last object that they are putting}? -- No \\ \emph{s1}: Which is the \object{last object that the person is putting}?  -- \textbf{NOT} \object{book}\end{tabular}                                               \\ \hline
\begin{tabular}[c]{@{}l@{}}\category{Choose} \\ \category{(Objects/ time)}\end{tabular}     & \begin{tabular}[c]{@{}l@{}}If ‘choose {[}\object{object1}{]} or {[}\object{object2}{]} {[}\object{indirect object}{]}’ is ‘\object{object1}’ \\ then {[}\object{object1}{]} equals {[}\object{indirect object}{]} should be ‘Yes’ \\ and {[}\object{object2}{]} equals {[}\object{indirect object}{]} should be ‘No’\end{tabular} & \begin{tabular}[c]{@{}l@{}}\emph{q }: Is the \object{doorway} or the \object{cup} the \object{first object they went behind}?  -- \object{doorway} \\ \emph{s1}: Is the \object{doorway} the \object{first object they went behind}? -- Yes\\ \emph{s2}: Is the \object{cup} the \object{first object they went behind}? -- No\end{tabular}                       \\ \cline{2-3}
& \begin{tabular}[c]{@{}l@{}}If ‘Does {[}\action{action1}{]} occur \temporal{before} or \temporal{after} {[}\action{action2}{]}’ is ‘\temporal{before}’ \\ then ‘Does {[}\action{action1}{]} occur \temporal{before} {[}\action{action2}{]}?’ should be ‘Yes’ \\ and ‘Does {[}\action{action1}{]} occur \temporal{after} {[}\action{action2}{]}?’ should be ‘No’\end{tabular}     & \begin{tabular}[c]{@{}l@{}}\emph{q }: Is the \object{person} \action{holding a cup} \temporal{before} or \temporal{after} \action{smiling at something}? -- \temporal{before} \\ \emph{s1}: Is the \object{person} \action{holding a cup} \temporal{before} \action{smiling at something}? -- Yes \\ \emph{s2}: Is the \object{person} \action{holding a cup} \temporal{after} \action{smiling at something}?  -- No\end{tabular} \\ \hline
\end{tabular}
}
\end{table*}

\begin{figure*}[t]
     \centering
     \includegraphics[width=\linewidth]{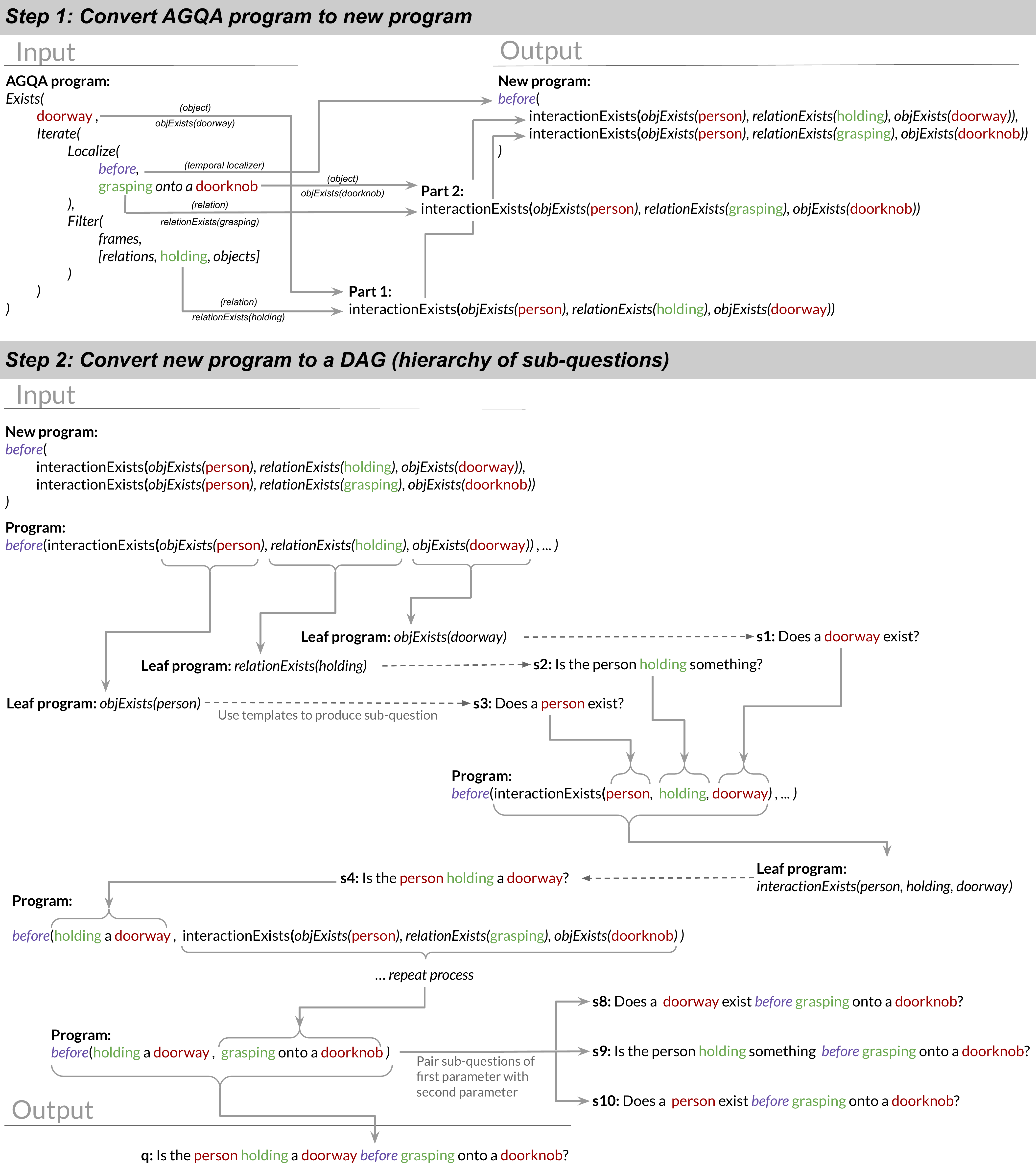}
     \caption{The figure shows the process of generating a question hierarchy using an AGQA program for the example AGQA question ``Is the person holding a doorway before grasping onto a doorknob?'' \textbf{Step 1:} We transform the AGQA program into a program representing the reasoning steps of the question. \textbf{Step 2:} We use Algorithm \ref{alg:question_creation} to generate the hierarchy of sub-questions from the new program.}
     \label{fig:detailed-process}
 \end{figure*}

\begin{figure*}[t]
     \centering
     \includegraphics[width=\linewidth]{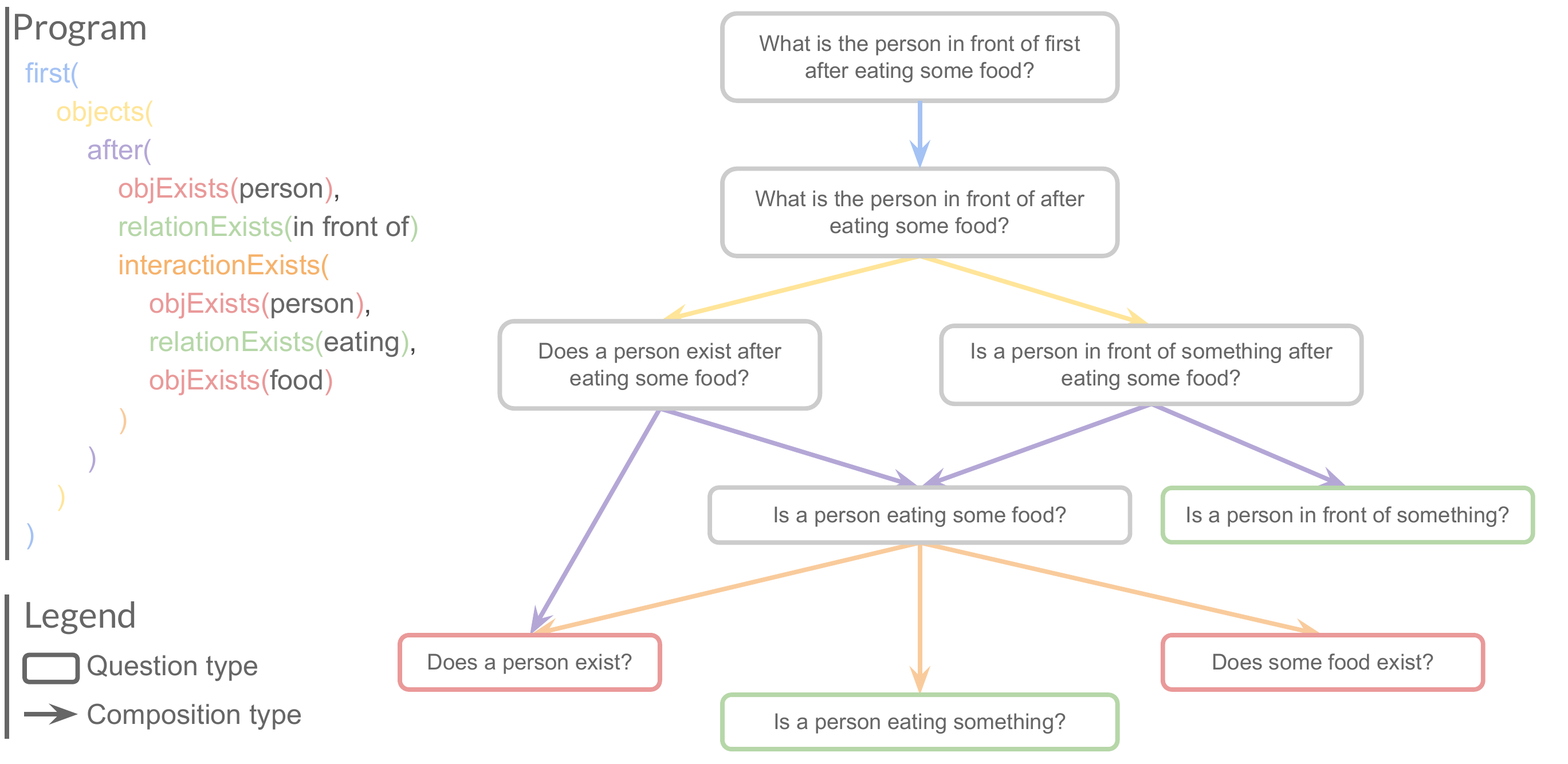}
     \caption{Example decomposition with corresponding program.}
     \label{fig:ex1}
 \end{figure*}

\paragraph{AGQA version.}
We use an updated version of AGQA\footnote{AGQA 2.0: \url{https://tinyurl.com/agqavideo}} that incorporates multiple improvements over the original dataset~\cite{GrundeMcLaughlin2021AGQA}. Throughout the paper we refer to this updated version as AGQA for simplicity. The most significant change is an updated balancing algorithm to further reduce linguistic biases. Some smaller improvements were motivated by minor errors in AGQA we discovered while ensure that AGQA-Decomp was internally consistent. 
 
\paragraph{AGQA program to subquestion hierarchy.}
In order to generate sub-question hierarchies, we first convert the original AGQA programs to a new program format. Each AGQA question type has a simple program template associated with it. To get compositional questions, AGQA makes this template more complex by introducing indirect references and temporal localization. As such, while forming the new programs, we firstly get the smaller programs for indirect references, if there are any, and continue by getting the temporal localization and the simple program associated with the basic template. We finally combine these to form the new program. 

For example, the AGQA type focusing on the existence of a relation between a person and an object has \texttt{Exists([\object{object}], Iterate(video, Filter(frame, [relations, [\relationship{relation}], objects])))} as its simple program. Using this structure of the AGQA original program template, we extract the \object{object} and \relationship{relation} for the new program. In this simple form, the corresponding new program is \texttt{interactionExists(objExists(person), relationExists([\relationship{relation}]), objExists([\object{object}]))}. We perform a similar process of translating between program types for temporal localization phrases (e.g. \texttt{Localize(before, \action{action})} translates into \texttt{before(\dots, \action{action} program)}). Step 1 of Figure~\ref{fig:detailed-process} visualizes the conversion of an AGQA program to the new program format. 

Upon converting AGQA programs into the new program format, we derive subquestion hierarchies from the new programs. Step 2 of Figure~\ref{fig:detailed-process} and Algorithm~\ref{alg:question_creation} illustrate the decomposition process for the newly generated program.

\paragraph{Use of the unbalanced AGQA dataset.}
Given question decompositions, our first strategy for obtaining ground-truth answers is to rely on the original AGQA annotations. This approach is not straightforward. After decomposing the questions in the balanced AGQA dataset, we can find sub-questions that are not present in the balanced dataset. 
If the parent is present in the balanced AGQA dataset, we are not guaranteed that the sub-question will also be present in the balanced version and cannot use the balanced dataset to derive its answer. Since the unbalanced AGQA dataset covers most of the possible questions (other than the newly added exists sub-questions), we rely on it to get answers for the question decompositions in the balanced version.

Specifically, we decompose $97$M questions from the unbalanced AGQA dataset, with each hierarchy having an average of $16.81$ sub-questions. This process produces $25.53$M unique new sub-questions. We determine the answers using different logical consistency rules for each video in the unbalanced dataset, which answers $87.92\%$ of the data. The question-answer pairs for a video from the decomposed unbalanced AGQA dataset are then used to answer our questions and our sub-questions for the same video. This process generates answers for in $90.23\%$ of the data in our balanced subquestion hierarchies. 

\paragraph{No exists questions.}\label{sec:no exist}
 Object Exists questions (e.g. ``Does a closet exist?'') are not a part of the original AGQA dataset. This is because the Action Genome scene graphs, which were used to generate the AGQA questions, only contain objects that an actor is interacting with~\cite{ji2020action}. Therefore, existing objects that are in the background, or that are extremely common (e.g. clothes, floor), are often not annotated. We can infer the ``yes'' answer through logical entailment (e.g. if the answer to the question ``Did they interact with $<$object$>$?'' is ``yes'', then its sub-question ``Does $<$object$>$ exist?'' must also be ``yes''). However, there is no way to use logical entailments to determine which objects do not exist.
 
 Therefore, we generate Object Exists questions answered ``no'' through two methods. First, we source human annotations for what objects do not exist within the video (see Human evaluation subsection). Then, we also include questions in which the object exists, but the temporal localization phrase contains an invalid action (``Does $<$object$>$ exist before they $<$invalid action$>$?''). These two methods generate $135K$ Object Exists questions with a ``no'' answer.

\paragraph{Limitations.}

There are limitations to our approach. First, this approach assumes AGQA answers to be ground truth. However, like all benchmarks, AGQA answers can be incorrect. These errors are described in more detail in their paper~\cite{GrundeMcLaughlin2021AGQA}.

Furthermore, not all questions in the hierarchies can be answered by the scene graph annotations AGQA uses as its basis for video representation~\cite{ji2020action,sigurdsson2016hollywood}. The AGQA scene graphs only annotate objects with which the actor is interacting, so they may miss existing objects in the background of the video or objects that are so generic that they often exist without annotations (e.g. ``floor'' or ``clothes''). The blacklisting of certain questions in AGQA also affected the subset of sub-questions in our decompositions that have associated AGQA answers.

\textbf{Societal impact.} Large curated datasets used to train vision models are known to contain biases, be it gender~\cite{hendricks2018women, zhao-etal-2017-men}, racial~\cite{buolamwini2018gender, zhao2021understanding} or geographic~\cite{wang2020revise}, or with problematic content~\cite{birhane2021multimodal}. Models trained on these datasets can then learn and propagate these biases to the real world, causing unintended harm. We note that AGQA-Decomp is primarily intended as a diagnostic dataset guiding model development and evaluation. A user leveraging AGQA-Decomp as training data should therefore recognize that models can propagate biases latent in the training data. Furthermore, as detailed in the limitation section, our automatic generalization process propagates error for ground-truth answers in the AGQA dataset, which can hurt real-world performance.

\begin{figure*}[t]
     \centering
     \includegraphics[width=\linewidth]{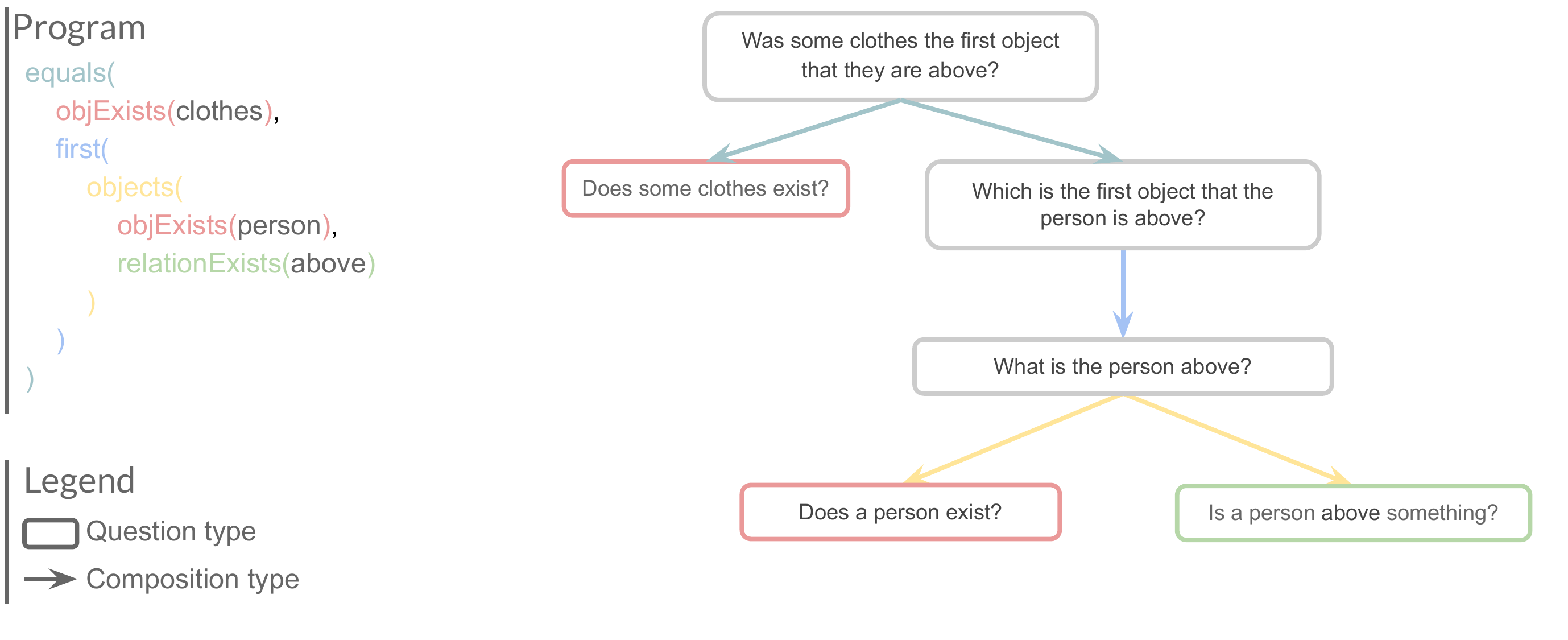}
     \caption{Example decomposition with corresponding program.}
     \label{fig:ex2}
 \end{figure*}
 
\begin{figure*}[t]
     \centering
     \includegraphics[width=\linewidth]{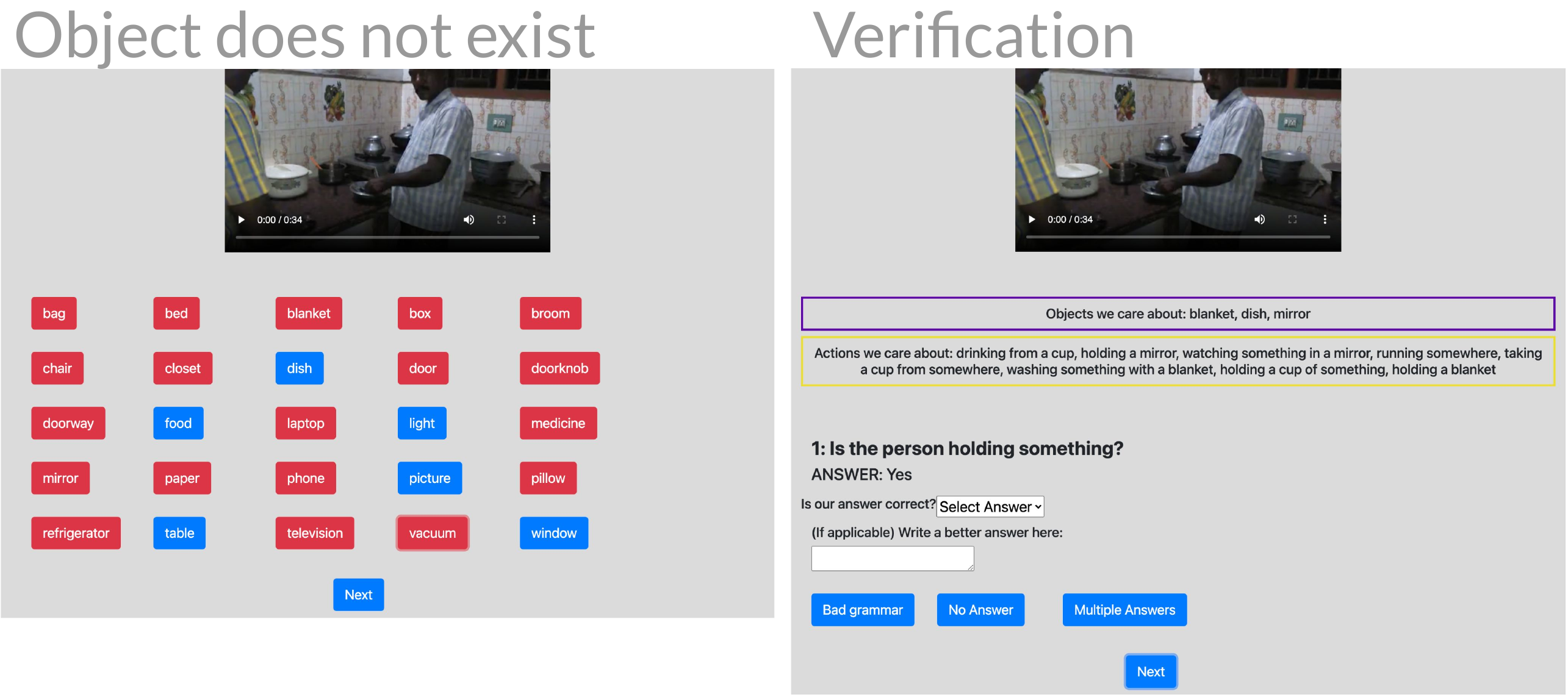}
     \caption{\textbf{Left: } The annotator views a video and a list of nearly all the objects in AMT. Annotators select the objects that do not appear in the video. \textbf{Right: } The annotator watches five videos that each appear with a question and an answer. Annotators indicate if that answer is Correct or Incorrect from a dropdown menu.}
     \label{fig:amt}
 \end{figure*}
 
\paragraph{Collecting more annotations.}
To identify missing objects from the scene graphs, we create an object labeling task. When we know that an object definitively doesn't exist, we can now answer questions that have the answer ``no'' (e.g.~``Did the person \relationship{touch} a \object{cup}?'' would be ``no'' if \object{cup} was not identified anywhere in the video.). We pay such that the equivalent hourly rate is $\$15$ per hour.

The question decomposition method cannot infer Object Exists questions with the answer ``no.'' Therefore, we run a study for human participants to mark which objects do not exist in the video. For a given video, participants must select the objects that do not appear in the video from a list of nearly all the objects in AGQA (See Figure~\ref{fig:amt}). We do not offer objects that nearly always exist (person, clothes, floor, hands, and hair).  We quality check by looking at whether they mark objects in the scene graph as present in the video. At the end of this process, we have the objects that do not exist for $88$ randomly selected videos. We were not able to to repeat this process on all videos due to monetary and time restrictions. We then take these objects and use the subquestion templates to generate questions. We use actions within the video to also generate questions with temporal localization phrases. This process generates $135K$ Object Exists questions with a ``No'' answer. 

\paragraph{Human evaluation.}
We evaluate the accuracy of sub-questions to find the error rate in each sub-question type.

Our answers in the question decompositions originate from the AGQA dataset as well as from logical entailments. Therefore, the errors that our human annotators mark in the questions originate from the AGQA dataset. The AGQA benchmark paper provides details about the source of these errors, including incorrect annotations, incorrect augmentations, inconsistent annotations, and human-AGQA definition mismatches~\cite{GrundeMcLaughlin2021AGQA}. We run the same validation task as the AGQA benchmark on at least $25$ questions per sub-question type. For all analysis, we take the the majority vote of $3$ annotators for each question. 

In this task annotators see a question, answer, and video. They are provided with a dropdown menu to mark the question as Correct or Incorrect. If they select Incorrect, we provide a space to write the correct answer. We also collect information on whether the question has bad grammar, multiple answers, or no possible answers. We check for the quality of responses with questions that we know to be answered incorrectly. Annotators mark $88.00\%$ of these incorrect questions as incorrect.

\begin{figure*}[t]
     \centering
     \includegraphics[width=\linewidth]{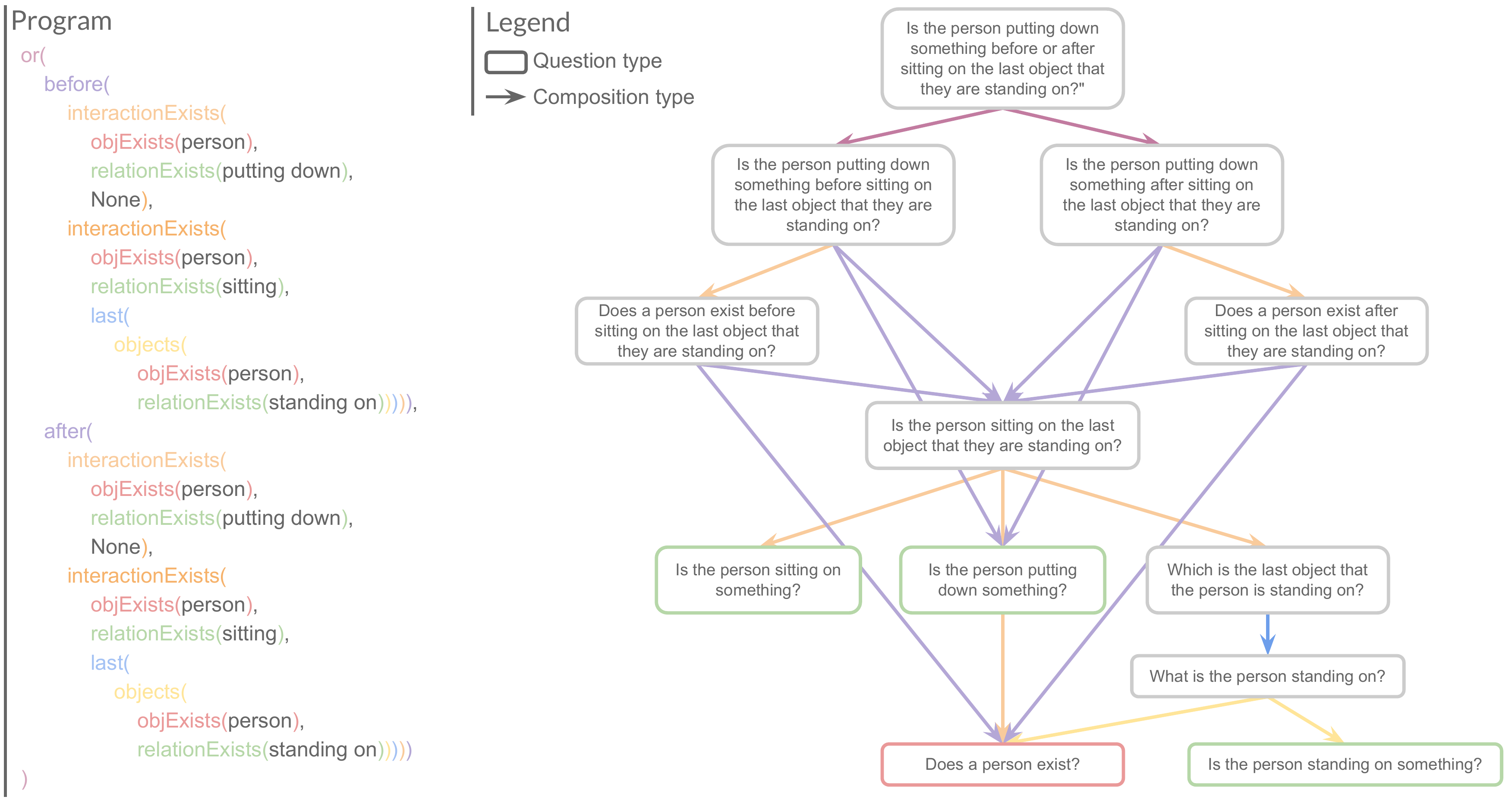}
     \caption{Example decomposition with corresponding program.}
     \label{fig:ex4}
 \end{figure*}

\subsection{Metrics}
In this section, we give additional details for our metrics. We first provide precise definitions for each metric. Afterwards, we give guidelines on how to interpret and compare values for each metric.

In Section $4$ of the main paper, we gave definitions for our metrics in plain English. We provide equations for each for further clarity. In all following definitions, let $f$ refer to the model we want to evaluate. Given an input video-question pair $(v,q)$, we set $\text{Acc}(v, q, f) = 1$ if $f$ made a correct prediction on this input and $0$ otherwise.

\paragraph{Compositional accuracy (\textsc{CA}):} We will begin with a formal definition of the metric's general form. Let $q$ be an arbitrary question and define $C_q$ to be the set of immediate sub-questions associated with $q$. To compute CA, we consider the set $Q_{CA}$ of all video-question pairs $(v,q)$ where $|C_q| > 0$ and $\text{Acc}(v, s, f) = 1$ for all $s \in C_q$. Then,
$$\text{CA}(f) = \frac{\sum_{(v, q) \in Q_{CA}}\text{Acc}(v, q, f)}{|Q_{CA}|}.$$
When we condition on question types, we compute the average on a subset of $Q_{CA}$ where the parent questions $q$ belong to a particular question type $p$ instead. The change is more complicated when we condition on composition rules, however. Let $t$ be the composition rule we are conditioning on. Then, for each question $q$, we change all instances of $C_q$ to $C_{q, t} = \{s \in C_q | (q, s, t) \in E_q\}$, where $E_q$ is the set of edges in the DAG associated with $q$. In plain English, we consider only the immediate sub-questions of $q$ related to it by the composition rule $t$.

\paragraph{Right for the wrong reasons (\textsc{RWR}):} The formulas for RWR are similar to those for CA. To compute RWR, we consider the set $Q_{RWR}$ of all video-question pairs $(v,q)$ where $|C_q| > 0$ and where there exists at least one $s \in C_q$ such that $\text{Acc}(v, s, f) = 0$. Then,
$$\text{RWR}(f) = \frac{\sum_{(v, q) \in Q_{RWR}}\text{Acc}(v, q, f)}{|Q_{RWR}|}.$$ We condition on question types and on composition rules using the exact method as for CA. 

To compute the more granular variant of RWR, RWR-n, we perform the same operations on the set $Q_{RWR-n}$ of all video-questions pairs $(v,q)$ where $|C_q| > 0$ and where the number of $s \in C_q$ such that $\text{Acc}(v, s, f) = 0$ is exactly $n$.

\paragraph{Delta:} Delta is defined as the difference between RWR and CA for a given model $f$:
$$\text{Delta}(f) = \text{RWR}(f) - \text{CA}(f).$$

\paragraph{Internal Consistency (\textsc{IC}):} We will begin with a formal definition of the metric's general form. Denote $\Phi$ as the set of all logical consistency rules. Let $\phi \in \Phi$ be any logical consistency rule, $(v, q)$ be any arbitrary video-question pair and $C_q$ be the set of immediate sub-questions associated with $q$. We then set $\phi(q, C_q, v, f) = 1$ if $f$'s predictions for $q$ and its sub-questions pass $\phi$'s consistency check, $0$ if it fails and $-1$ if the check cannot be applied to the given set of question-answer pairs. In order to compute internal consistency for a given logical consistency rule $\phi \in \Phi$, denoted $IC_\phi$, we consider the set $Q^\phi_{IC}$ of all video-question pairs $(v, q)$ such that $\phi(q, C_q, v, f) \neq -1$. We then define $$IC_\phi(f) = \frac{\displaystyle \sum_{(v,q) \in Q^\phi_{IC}}\phi(c, C_q, v, f)}{|Q^\phi_{IC}|}.$$ The overall $IC$ metric is then defined as $$IC(f) = \frac{\sum_{\phi \in \Phi}IC_\phi(f)}{|\Phi|}.$$ If any $IC_\phi(f)$ is undefined due to $|Q^\phi_{IC}| = 0$, we also treat $IC(f)$ as undefined.

In order to condition on a particular composition rule $t$ for $IC$, we simply perform the same operations using the set of logical consistency rules $\Phi_t$ applicable to $t$ instead of the general set $\Phi$. Conditioning on a specific parent question type $p$ is similar, but more complicated. As before, we restrict our attention to the set of logical consistency rules $\Phi_p$ applicable to the parent question type $p$. However, we further focus on subsets of $Q^\phi_{IC}$ where the parent questions $q$ belong to the question type $p$. We finally note that the compositions logical consistency rules associated with a particular composition rule check can overlap. This can result in double counting of failed consistency checks when computing IC values for composition rules or parent question types.

\begin{figure*}[t]
     \centering
     \includegraphics[width=\linewidth]{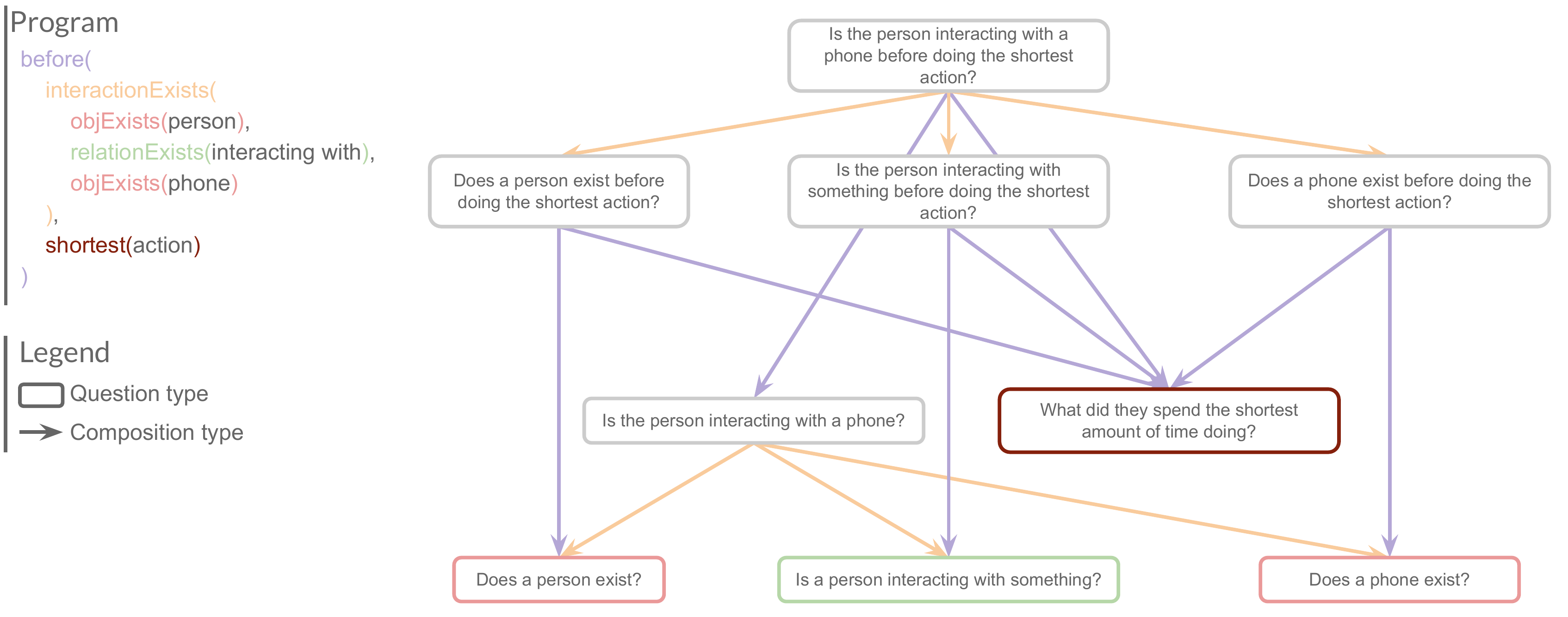}
     \caption{Example decomposition with its corresponding program.}
     \label{fig:ex3}
 \end{figure*}
 
\paragraph{Accuracy:} We compute accuracy per question type and normalize across answers to obtain an aggregate value. Consider any question type $t$ and let $A_t$ be the set of ground-truth answers associated with questions of type $t$. Referring to $Q_{t, a}$ as the set of video-question pairs $(v, q)$ where $q$ is of type $t$ and for which $a$ is the ground-truth answer, we formally define
$$\text{Accuracy}(f,t) = \frac{\displaystyle \sum_{a \in A_t}\frac{\sum_{(v,q) \in Q_{t,a}}\text{Acc}(v, q, f)}{|Q_{t,a}|}}{|A_t|}.$$

\paragraph{Interpreting Values for Metrics.}
We expect a model that reasons compositionally to have high values for the Accuracy, CA, and IC metrics and to have low values for the RWR metric. Given that we expect a model to perform poorer on parent questions when it answers at least one sub-question incorrectly, we also expect a model that reasons compositionally to obtain negative Delta values. In other words, we expect RWR to always be lower than CA.

In the event when a model obtains desirable values for each metric, it is fruitful to perform more granular analysis, inspecting model performances for the various RWR-n metrics, individual composition rules and ground-truth answers in addition to qualitative analysis.

\begin{figure*}[t]
     \centering
     \includegraphics[width=\linewidth]{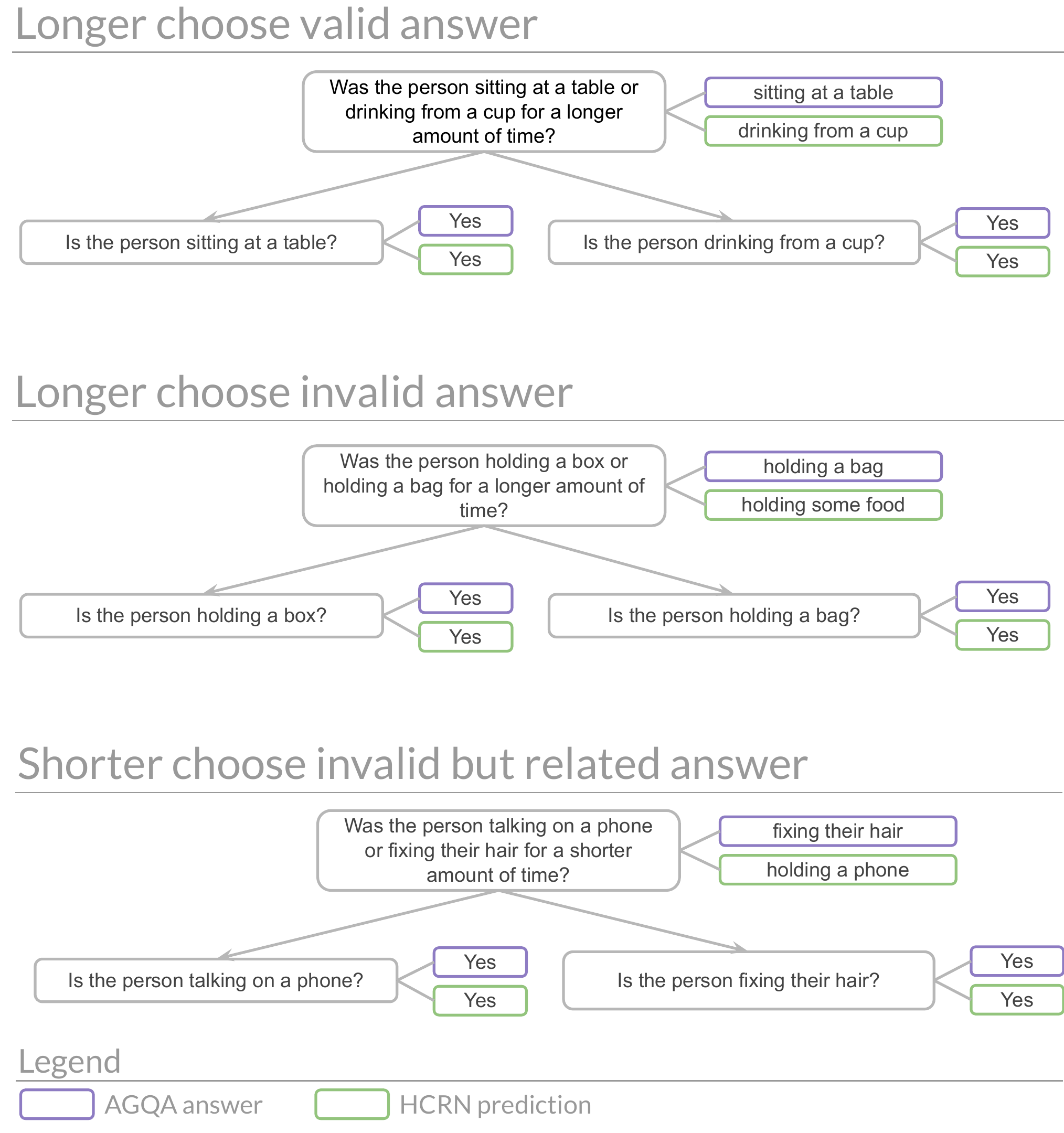}
     \caption{We present example compositions where HCRN answers all children correctly but answers the parent incorrectly. \textbf{Top:} HCRN picks a valid but inaccurate option. \textbf{Center:} HCRN gives an unrelated response. \textbf{Bottom:} HCRN produces an invalid but relevant answer.}
     \label{fig:qualitative}
 \end{figure*}

\begin{figure*}[t]
     \centering
     \includegraphics[width=\linewidth]{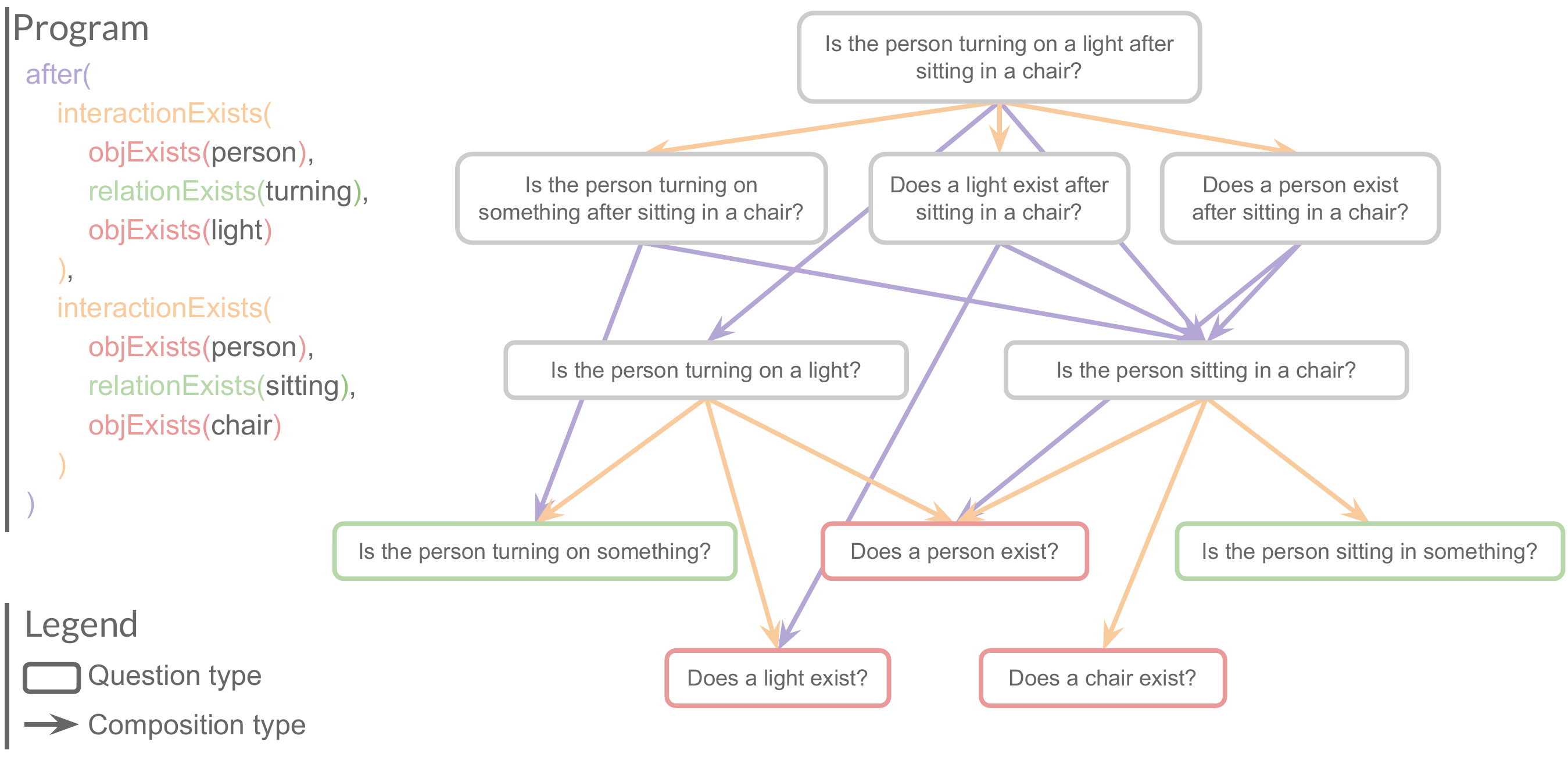}
     \caption{Example decomposition with corresponding program.}
     \label{fig:ex5}
 \end{figure*}

\subsection{Experiments}
In this section, we first describe the question types that we ignored during evaluation due to poor human validation scores and then detail how we trained and evaluated models. Afterwards, we perform an experiment exploring the use of AGQA-Decomp as data augmentation and provide additional analyses for the Most Likely baseline. We finally give examples of error modes that appeared during qualitative analysis.

\paragraph{Banned Question types.}

When evaluating model performance on the questions, we ignore questions of types that did not achieve at least a $70\%$ human validation score. The following types did not achieve this threshold.

\begin{itemize}
  \item \textbf{Action Temporal Localization}: This question type contains open answer questions for action recognition such as ``What were they doing after walking through the doorway?'' Human annotators marked $55.00\%$ of questions of this type as correct. 
  \item \textbf{Object}: This question type contains open answer questions for objects such as ``What were they opening?''. It also includes such questions when they have a temporal localization phrase. Human annotators marked $62.16\%$ of questions of this type as correct.
\end{itemize}

Future work could address limitations in the AGQA dataset in order to improve the accuracy of questions or create a more accurate subset of questions. This new version could then be used to evaluate all types of questions. 

\begin{table*}[]
\caption{We present HCRN, HME and PSAC performances on the \textbf{RWR-n} metrics, where $n$ represents the exact number of incorrectly answered sub-questions for a composition, while conditioning on parent question types. Models are frequently accurate on parent questions even when answering simpler sub-questions incorrectly. For \category{Equals} and particularly \category{Interaction Temporal Localization} questions, \textbf{RWR-n} values largely outperform \textbf{CA} scores}
\label{tab:count_treewide}
\centering
\resizebox{\linewidth}{!}{
\begin{tabular}{l rrrrrr rrrrrr rrrrr}
                                  & \multicolumn{6}{c}{HCRN}                        & \multicolumn{6}{c}{HME}                         & \multicolumn{5}{c}{PSAC}                        \\ \cline{2-6} \cline{8-12} \cline{14-18} 
Parent Type                       & 
RWR-1 & RWR-2 & RWR-3 & RWR-4 & RWR-5 & &
RWR-1 & RWR-2 & RWR-3 & RWR-4 & RWR-5 & &
RWR-1 & RWR-2 & RWR-3 & RWR-4 & RWR-5 \\ \hline\hline
\category{Object Exists}                     & 
N/A     & N/A     & N/A     & N/A     & N/A     & &
N/A     & N/A     & N/A     & N/A     & N/A     & &
N/A     & N/A     & N/A     & N/A     & N/A     \\
\category{Relation Exists}                   & 
16.67   & N/A     & N/A     & N/A     & N/A     & &
N/A     & N/A     & N/A     & N/A     & N/A     & &
20.22   & N/A     & N/A     & N/A     & N/A     \\
\category{Interaction}                       & 
44.26 &	29.40 &	19.55 &    N/A     & N/A     & &
26.63 &	12.15 & N/A     & N/A     & N/A     & &
63.19 &	49.14 &	26.78    & N/A     & N/A     \\
\category{Interaction Temporal Loc.} & 
49.23 &	55.34 &	56.26 &	39.65 &	6.25    & &
89.05 &	93.92 &	70.33 &	71.69 &	4.35     & &
29.37 &	58.69 &	65.44 &	28.15 &	9.12   \\
\category{Exists Temporal Loc.}      & 
67.93 &	21.98   & N/A     & N/A     & N/A     & &
2.25 &	1.83    & N/A     & N/A     & N/A     & &
30.23 &	4.52    & N/A     & N/A     & N/A     \\
\category{First/Last}                        & 
N/A     & N/A     & N/A     & N/A     & N/A     & &
N/A     & N/A     & N/A     & N/A     & N/A     & &
N/A     & N/A     & N/A     & N/A     & N/A     \\
\category{Longest/Shortest Action}           & 
N/A     & N/A     & N/A     & N/A     & N/A     & &
N/A     & N/A     & N/A     & N/A     & N/A     & &
N/A     & N/A     & N/A     & N/A     & N/A     \\
\category{Conjunction}                       & 
48.17   & 31.21   & N/A     & N/A     & N/A     & &
47.42   & 27.85   & N/A     & N/A     & N/A     & &
46.17   & 30.60   & N/A     & N/A     & N/A     \\
\category{Choose}                            & 
47.51 &	41.12   & N/A     & N/A     & N/A     & &
48.57 &	39.74   & N/A     & N/A     & N/A     & &
48.24 &	47.32   & N/A     & N/A     & N/A     \\
\category{Equals}                            & 
52.10 &	50.20   & N/A     & N/A     & N/A     & &
47.08 &	47.54   & N/A     & N/A     & N/A     & &
51.19 &	47.69   & N/A     & N/A     & N/A     \\ \hline
Overall                           & 
55.59 &	32.24 &	47.31 &	39.65 &	6.25    & &
30.05 &	13.43 &	70.33 &	71.69 &	4.35     & &
40.75 &	27.73 &	57.47 &	28.15 &	9.12  
\end{tabular}
}
\end{table*}
\begin{table*}[]
\caption{We present HCRN, HME and PSAC performances on the \textbf{RWR-n} metrics, where $n$ represents the exact number of incorrectly answered sub-questions for a composition, while conditioning on composition rules between questions and their sub-questions. Models are frequently accurate on parent questions even when answering simpler sub-questions incorrectly. \textbf{RWR-1} and \textbf{RWR-2} scores reveal problematic reasoning for \category{And} and \category{Xor} compositions respectively for HME and PSAC.}
\label{tab:count_parentchild}
\centering
\resizebox{\linewidth}{!}{
\begin{tabular}{l rrrr rrrr rrr}
                & \multicolumn{4}{c}{HCRN}    & \multicolumn{4}{c}{HME}     & \multicolumn{3}{c}{PSAC}    \\ \cline{2-4} \cline{6-8} \cline{10-12} 
Compostion Type & 
RWR-1 & RWR-2 & RWR-3 & &
RWR-1 & RWR-2 & RWR-3 & &
RWR-1 & RWR-2 & RWR-3 \\ \hline\hline
\category{Interaction}     & 
50.67 &	42.62 &	17.86   & &
48.80 &	76.75 &	11.18    & &
50.93 &	63.22 &	23.81   \\
\category{First}           & 
N/A     & N/A     & N/A     & &
N/A     & N/A     & N/A     & &
N/A     & N/A     & N/A     \\
\category{Last}            & 
N/A     & N/A     & N/A     & &
N/A     & N/A     & N/A     & &
N/A     & N/A     & N/A     \\
\category{Equals}          & 
52.10 &	50.20   & N/A     & &
47.08 &	47.54   & N/A     & &
51.19 &	47.69   & N/A     \\
\category{And}             & 
48.35   & 15.40   & N/A     & &
79.38   & 6.32    & N/A     & &
80.11   & 10.61   & N/A     \\
\category{Xor}             & 
48.03   & 51.99   & N/A     & &
24.95   & 86.73   & N/A     & &
22.54   & 82.81   & N/A     \\
\category{Choose}          & 
47.72 &	42.20    & N/A     & &
48.63 &	36.76   & N/A     & &
48.47 &	47.59   & N/A     \\
\category{Longer Choose}   & 
36.84 &	40.32    & N/A     & &
44.01 &	39.58    & N/A     & &
40.66 &	41.99   & N/A     \\
\category{Shorter Choose}  & 
37.19 &	36.52   & N/A     & &
43.99 &	40.84   & N/A     & &
41.04 &	42.56   & N/A     \\
\category{After}           & 
61.24 &	33.33   & N/A     & &
17.92 &	24.39  & N/A     & &
37.08 &	15.52   & N/A     \\
\category{Before}          & 
65.16 &	36.90   & N/A     & &
17.40 &	23.90   & N/A     & &
34.81 &	15.77   & N/A     \\
\category{While}           & 
66.01 &	21.34   & N/A     & &
10.09 &	8.94    & N/A     & &
32.36 &	10.20   & N/A     \\
\category{Between}         & 
33.14   & 5.98    & N/A     & &
77.83   & 1.34    & N/A     & &
41.01   & 4.69    & N/A     \\ \hline
Overall         & 
55.12 &	32.97 &	17.86   & &
36.73 &	23.56 &	11.18    & &
42.84 &	31.63 &	23.81  
\end{tabular}
}
\end{table*}
\begin{table}[]
\caption{We present internal consistency (\textbf{IC}) scores for individual logical consistency rules for HCRN, HME, PSAC and the Most-Likely baseline. Logical consistency rules being followed by ``Yes'' or ``No'' indicates that the parent question either is or implied to be ``Yes'' or ``No''. For \category{Choose} questions, ``Object'' and ``Temporal'' denote whether the parent is an object or ``before'' or ``after.'' Models frequently achieve low values when the parent is ``Yes'' and are particularly inconsistent for \category{Choose} consistency rules.}
\label{tab:ic_splits}
\centering
\resizebox{\columnwidth}{!}{
\begin{tabular}{l r rr rrr}
                  Consistency & Parent & & \multicolumn{4}{c}{IC}                      \\ \cline{4-7} 
Check & Answer
& & HCRN     & HME      & PSAC    & Most-Likely \\ \hline\hline
\category{Interaction} & Yes & & 75.62 &	26.69  & 0.00  & 100.00    \\
 & No & & 75.23 &	96.48 &	56.64 & N/A         \\ \hline
\category{Equals} & Yes &     & 6.04 &	2.74 &	3.31 &	6.84       \\
 & No  &     & 50.17 &	83.96 &	75.21  & 0.00         \\ \hline
\category{And} & Yes  &       & 69.10  & 30.05  & 5.22  & 0.00         \\
 & No   &       & 78.98  & 98.2  & 90.89 & 0.00      \\\hline
\category{Xor} & Yes  &      & 4.06   & 1.54  & 9.38 & N/A         \\
 & No    &      & 50.03  & 87.59  & 89.64 & 100.00    \\ \hline
\category{Choose} & Object &   & 6.59 &	0.76 &	14.80 & N/A         \\
 & Temporal & & 4.92 &	0.54 &	9.56   & 0.00      \\ \hline
\category{After} & Yes  &     & 40.65 &	40.22 &	42.49 & 100.00    \\
 & No  &      & 54.17 &	98.89 &	70.15 & N/A         \\ \hline
\category{Before} & Yes  &    & 38.64 &	41.92 &	43.48 & 100.00    \\
 & No &      & 49.28  & 98.94  & 70.86 & N/A         \\ \hline
\category{While} & Yes  &    & 42.64 &	33.85 &	44.76 & 100.00    \\
 & No   &     & 59.95 & 98.68 & 80.02 & N/A         \\ \hline
\category{Between} & Yes  &   & 77.30 &	37.72 &	73.10 & 100.00    \\
 & No  &    & 73.82  & 98.76  & 89.97 & N/A         \\ \hline
Overall      &   &  & 47.62 &	54.31 &	48.30 & N/A    
\end{tabular}
}
\end{table}
\begin{table}[]
\caption{We report accuracy per ground-truth answer for each binary question type expecting "Yes" or "No" answers for HCRN, HME, PSAC and the Most-Likely baseline. Models frequently perform well on one ground-truth answer at the expense of the other. HME particularly is biased towards "No" for all question types except \category{Object Exists}.}
\label{tab:accuracy_splits}
\centering
\resizebox{\columnwidth}{!}{
\begin{tabular}{l r rr rrr}
                                     Question   & Ground & & \multicolumn{4}{c}{Accuracy}        \\ \cline{4-7} 
Type & Truth  &          & HCRN  & HME   & PSAC  & Most-Likely \\ \hline
\category{Object Exists} & Yes &                    & 44.39 & 93.47 & 3.38  & 100.00      \\
 & No       &               & 49.70 & 0.00  & 86.67 & 0.00        \\ \hline
\category{Relation Exists} & Yes    &              & 48.43 & 3.29  & 68.85 & 100.00      \\
 & No      &              & 55.84 & 99.11 & 4.02  & 0.00        \\ \hline
\category{Interaction} & Yes    &                   & 39.78 & 9.16  & 40.91 & 100.00      \\
 & No       &                 & 53.65 & 91.98 & 83.76 & 0.00        \\ \hline
 \category{Interaction Temporal} & Yes & & 57.15 & 3.60  & 40.82 & 100.00      \\
 \category{Loc.} & No &  & 41.91 & 97.24 & 49.58 & 0.00        \\ \hline
\category{Exists Temporal} & Yes  &    & 59.42 & 1.32  & 28.58 & 100.00      \\
\category{Loc.} & No    &   & 36.21 & 98.06 & 78.46 & 0.00        \\ \hline
\category{Conjunction} & Yes    &                   & 41.32 & 1.33  & 7.07  & 0.00        \\
 & No         &               & 57.88 & 98.82 & 92.94 & 100.00      \\ \hline
\category{Equals} & Yes      &                      & 41.91 & 1.88  & 19.80 & 100.00      \\
 & No     &                        & 59.15 & 98.28 & 80.05 & 0.00        \\ \hline
\end{tabular}
}
\end{table}

\paragraph{Training Details.}
Upon running initial experiments with the default configurations of HCRN, HME and PSAC's respective repositories, we found that HCRN and PSAC overfit our data. As such, we performed hyperparameter searches for learning rate and weight decay parameters and additionally incorporated new dropout layers for each model to improve regularization. HCRN's best performing run was trained with a learning rate of $0.00016$, a weight decay of $0.0005$, a dropout probability of $0.15$ and a batch size of $32$. HME's best performing run remained the default configuration with a learning rate of $0.001$, a weight decay of $0.0$, no new dropout layers and a batch size of $32$. PSAC, finally, was trained with a learning rate of $0.003$, a weight decay of $5 * 10^{-6}$, a dropout probability of $0.15$ and a batch size of $32$. We trained HCRN for $5$ epochs (where each epoch performs $18$ validation loops), HME for $32000$ update steps (corresponding to $40$ validation loops) and PSAC for $23$ epochs. We began terminated training after the validation accuracy of each model had plateaued. HCRN, HME and PSAC achieved best validation accuracies of $46.48\%$, $42.492\%$ and $43.69\%$ at the point of evaluation.

\begin{table}[]
\caption{We present the accuracy the best performing HCRN, HME and PSAC runs obtain when trained on the AGQA or the AGQA-Decomp balanced training sets. Models trained on AGQA-Decomp outperform those trained on AGQA, implying that our DAGs may potentially be useful sources of data augmentation. Accuracy for this table is the standard definition of accuracy.}
\label{tab:data_augmentation}
\centering
\resizebox{\columnwidth}{!}{
\begin{tabular}{l rrr}
                 & \multicolumn{3}{c}{AGQA Accuracy} \\
Training dataset & HCRN      & HME       & PSAC      \\
\hline
AGQA             & 42.11     & 39.89     & 40.18     \\
AGQA-Decomp      & \textbf{43.10}     & 38.96     & 39.75 \\
\hline
\end{tabular}
}
\end{table}
\begin{table}[]
\caption{We report compositional accuracy (\textbf{CA}), right for the wrong reasons (\textbf{RWR}), delta (\textbf{RWR-CA}) and internal consistency (\textbf{IC}) metrics for the Most-Likely baseline with respect to question types. We find that whatever good performance the Most-Likely baseline achieves is within narrow slices of the dataset. N/A values under the IC column indicate that the model has no valid datapoints for at least one logical consistency rule for that question type.}
\label{tab:mc_treewide}
\centering
\resizebox{\columnwidth}{!}{
\begin{tabular}{lrrrr}
                                  & CA          & RWR         & Delta       & IC          \\ \hline
Question Type                     & Most-Likely & Most-Likely & Most-Likely & Most-Likely \\ \hline\hline
\category{Object Exists}                    & N/A         & N/A         & N/A         & N/A         \\
\category{Relation Exists}                   & 100.00      & N/A         & N/A         & N/A      \\
\category{Interaction}                       & 79.00       & 87.61       & 8.61        & N/A      \\
\category{Interaction Temporal Loc.} & 57.96       & 1.29        & -56.67      & N/A      \\
\category{Exists Temporal Loc.}      & 98.79       & 97.58       & -1.21       & N/A      \\
\category{First/Last}                        & N/A         & N/A         & N/A         & N/A         \\
\category{Longest/Shortest Action}           & N/A         & N/A         & N/A         & N/A         \\
\category{Conjunction}                       & 24.35       & 62.67       & 38.32       & N/A       \\
\category{Choose}                            & 6.02        & 24.48       & 18.46       & N/A       \\
\category{Equals}                            & 46.66       & 53.56       & 6.90       & 3.42       \\ \hline
Overall                           & 80.06       & 37.97       & -42.09      & N/A      
\end{tabular}
}
\end{table}
\begin{table}[]
\caption{We report compositional accuracy (\textbf{CA}), right for the wrong reasons (\textbf{RWR}), delta (\textbf{RWR-CA}) and internal consistency (\textbf{IC}) metrics for the Most Likely baseline with respect to composition rules. We find that whatever good performance the Most-Likely baseline achieves is within narrow slices of the dataset, such as the case when parent and child questions are answered ``No'' and ``Yes'' respectively for \category{Xor}. N/A values under the IC column indicate that the model has no valid datapoints for at least one logical consistency rule.}
\label{tab:mc_parentchild}
\centering
\resizebox{\columnwidth}{!}{
\begin{tabular}{lrrrr}
                 & CA          & RWR         & Delta       & IC          \\ \hline 
Composition Type & Most-Likely & Most-Likely & Most-Likely & Most-Likely \\ \hline\hline
\category{Interaction}      & 64.07       & 48.91       & -15.16      & N/A      \\
\category{First}            & N/A         & N/A         & N/A         & N/A         \\
\category{Last}             & N/A         & N/A         & N/A         & N/A         \\
\category{Equals}           & 46.66       & 53.56       & 6.90       & 3.42       \\
\category{And}              & 0.00        & 100.00      & 100.00      & 0.00        \\
\category{Xor}              & 100.00      & 40.39       & -59.61      & N/A      \\
\category{Choose}           & 18.52        & 24.48       & 5.96       & N/A       \\
\category{Longer Choose}    & 5.73        & N/A         & N/A         & N/A         \\
\category{Shorter Choose}   & 6.22        & N/A         & N/A         & N/A         \\
\category{After}            & 79.91       & 53.15       & -26.76       & N/A      \\
\category{Before}           & 80.51       & 54.22       & -26.30       & N/A      \\
\category{While}            & 93.32       & 52.05       & -41.26      & N/A      \\
\category{Between}          & 99.03       & 0.00        & -99.03      & N/A      \\ \hline
Overall          & 75.60       & 37.70       & -37.90      & N/A      
\end{tabular}
}
\end{table}

\paragraph{Using AGQA-Decomp as Data Augmentation} Another intuitive application of AGQA-Decomp is data augmentation for the original AGQA dataset \cite{GrundeMcLaughlin2021AGQA}. The training data we used for our main evaluation is a version of the AGQA balanced dataset augmented with a balanced subset of questions taken from our DAGs. We can therefore investigate whether our trained models' performances are better than those trained on the standard AGQA dataset. We compare the accuracies of the best performing runs for both sets of models and find that using the AGQA-Decomp sub-question data naively as data augmentation does not result in a clear improvement. HCRN trained on AGQA-Decomp outperforms its counterpart trained on AGQA by $1\%$, while HME, for instance, underperforms by $1\%$ (Table \ref{tab:data_augmentation}). One possible reason for the lack of improvement is our use of sub-questions naively as more data. Future work may devise data augmentation schemes that go beyond this naive approach and leverage the structure provided by entire hierarchies for potentially better performance. 

\paragraph{Further Comparisons with Most-Likely.}
We will provide further results and analyses involving the Most-Likely baseline in this section. The Most Likely baseline represents a model that relies primarily on linguistic biases, outputting the most likely answer for each basic question type. We will begin with a discussion of the Most-Likely baseline's IC results and then investigate individual question types and composition rules. 

\noindent\textbf{Performance on the IC metric:} 
The Most-Likely baseline, on one hand, is perfectly consistent for one half of logical consistency rules, primarily the rules where the parent is answered ``yes'' and all child answers are also propagated to be ``yes''. On the other hand, it has no valid data points for the other half of the rules (Table \ref{tab:ic_splits}). This is due to the fact that the Most-Likely baseline outputs the most common answer for each question type, severely restricting the parent-child answer distributions for each composition. Our overall IC metric avoids treating such biased models as highly consistent by performing a macro average of the consistency scores for each logical consistency rule associated with a question type or composition rule. For every single composition rule but \category{Equals} and \category{And}, the Most-Likely baseline has undefined performances on a logical consistency rule, which results in the undefined IC values on Tables \ref{tab:mc_treewide} and \ref{tab:mc_parentchild}. On the other hand, the Most-Likely baseline's performance is consistently poor for logical consistency rules associated with \category{Equals} and \category{And} compositions.

\noindent\textbf{Performance on Choose and Equals:} 
For the \category{Choose} question type, a category that contains a large set of possible answers, the Most-Likely baseline's performance is predictably poor with a CA score of $6.02\%$ (Table \ref{tab:mc_treewide}). The model only has valid datapoints for consistency checks on \category{Choose} compositions requiring choosing whether an event occurred before or after another. For this composition rule, the model is inconsistent for each case, as the child questions, which belong to the same question type, must be answered differently. Performance on the \category{Equals} category is also poor, with the model being self-consistent only $6.84\%$ of the time when the parent is answered ``yes'' (Table \ref{tab:ic_splits}).

\noindent\textbf{Performance on Conjunction:}
For \category{Conjunction} questions, the Most-Likely baseline is biased towards ``no'' answers while it is biased towards ``yes'' answers for sub-questions to \category{Conjunction} questions. As the \category{Xor} composition is always accurate for this answer distribution, the Most-Likely baseline obtains perfect CA score. Similarly, since the \category{And} composition is always inaccurate for this answer distribution, the model obtains $0.00\%$ for CA. These extreme CA scores, the model's undefined or poor IC values, as well as the high RWR score for \category{And} (Table \ref{tab:mc_parentchild}) collectively indicate incorrect reasoning.

\noindent\textbf{Performance on Temporal Reasoning:}
For temporal reasoning question types, such as \category{Exists Temporal Localization} and \category{Interaction Temporal Localization}, and their constituent composition rules (\category{After}, \category{Before}, \category{While}, \category{Between}), any good performance can be explained by the fact that the Most-Likely baseline answers only ``yes'' to both parent questions and its children. For these instances, the model is perfectly consistent. The IC scores being undefined on Tables \ref{tab:mc_treewide} and \ref{tab:mc_parentchild} alert that the model does not reason compositionally yet again.

\paragraph{Qualitative Examples.}
In this section, we provide example illustrations of error modes we observed when models answered all immediate sub-questions questions correctly but answered the parent question incorrectly for the composition rule in which models achieved the worst performance: Longer and Shorter Choose. Figure \ref{fig:qualitative} displays three error categories: one category where the model chooses the wrong option, one category where the model makes a semantically relevant prediction that is not given as an option and another category where the model makes a wholly irrelevant prediction.

\paragraph{Differences from CVPR Camera Ready Version} For the camera ready version, the scripts we used to compute internal consistency had two distinct bugs. Firstly, the script computing correlations between internal consistency and accuracy across DAGs contained a bug that produced values that were less negative than they should have been. We have amended Section 5.6, Figure 3 and the discussion at the end of Section 1 to reflect the correct values. Secondly, the script computing internal consistency values for Tables \ref{tab:treewide}, \ref{tab:parentchild}, \ref{tab:ic_splits}, \ref{tab:mc_treewide} and \ref{tab:mc_parentchild} contained a bug overestimating model performance when the parent question was answered ``no.'' We have amended the values on the tables and altered Sections 5.4 and 5.5, and the Most-Likely baseline discussion in the Supplementary to quote the corrected values.

\subsection{Future work}

While our analyses are limited to the AGQA benchmark, our decomposition structure can nonetheless facilitate multiple future contributions. 

\noindent\textbf{Consistency as a training loss} Following in the path laid out by recent work~\cite{selvaraju2020squinting,gokhale2020vqa,yuan2021perception}, consistency can be operationalized as an additional training signal to encourage models to behave compositionally. The proliferation recent large language models~\cite{wu2021polyjuice} can be prompted to produce consistent training data augmentations for smaller models. 

\noindent\textbf{Interactive model inspection:} Although the metrics that we propose each facilitate analyses across the entire dataset, they are motivated by how we expect models that reason compositionally should behave on individual examples. This makes the exploration of question DAGs as a tool for the interactive analysis of model behavior ~\cite{wu2019errudite,wu2021polyjuice} a fruitful direction. 

\noindent\textbf{Explanations through question decompositions:} Furthermore, model answers to question hierarchies can be used as justifications of model predictions, similar to past work on natural language rationalizations ~\cite{hendricks2016generating, hendricks2018grounding}, with each answer representing model behavior in intermediate reasoning steps ~\cite{dalvi2021explaining}. Internal consistency can similarly help determine whether to trust and rely on models.

This paper outlines several evaluation methods using a decomposition of AGQA questions. This application of a question decomposition structure already provides fruitful insights on model performance. The structure of AGQA-Decomp hierarchies can further provide both flexibility and nuance to evaluation outside of the use case explored here. We encourage future work to expand this structure to other benchmarks and to create novel evaluation methods.

\end{document}